\documentclass[10pt,journal,compsoc]{IEEEtran}

\usepackage{bbding}
\usepackage{tabularx}
\usepackage{graphicx}
\usepackage{booktabs}
\usepackage{makecell}
\usepackage{multirow}
\usepackage{float}
\usepackage{longtable}
\usepackage{ltxtable}
\usepackage{cite}
\usepackage{amsmath, amssymb, amsfonts}
\usepackage{algorithmicx}
\usepackage{textcomp}
\usepackage{xcolor} 
\usepackage{hyperref}
\usepackage{comment}
\usepackage{longtable}
\usepackage{booktabs}
\usepackage{graphicx}

\ifCLASSINFOpdf
\else
\fi
\begin{document}
%
\title{A Survey of Multimodal Composite Editing and Retrieval}

\author{Suyan Li,  Fuxiang Huang,  and Lei Zhang, \textit{Senior Member, IEEE}
\IEEEcompsocitemizethanks{
\IEEEcompsocthanksitem This work was partially supported by National Key R\&D Program of China (2021YFB3100800),  National Natural Science Fund of China (62271090),  Chongqing Natural Science Fund (cstc2021jcyj-jqX0023). This work is also supported by Huawei computational power of Chongqing Artificial Intelligence Innovation Center. (Corresponding author: Lei Zhang)
\IEEEcompsocthanksitem Suyan Li is with the Chongqing Key Laboratory of Bio-perception and Multimodal Intelligent Information Processing, Chongqing University,  Chongqing 400044,  China.
(E-mail: suyanli220@gmail.com)
\IEEEcompsocthanksitem Fuxiang Huang is with the Chongqing Key Laboratory of Bio-perception and Multimodal Intelligent Information Processing, Chongqing University,  Chongqing 400044, China, and the Hong Kong University of Science and Technology, Hong Kong.
(E-mail: huangfuxiang@cqu.edu.cn)
\IEEEcompsocthanksitem Lei Zhang is with the Chongqing Key Laboratory of Bio-perception and Multimodal Intelligent Information Processing, and the School of Microelectronics and Communication Engineering, Chongqing University, Chongqing 400044,  China.
(E-mail: leizhang@cqu.edu.cn)
}
}

\markboth{Journal of \LaTeX\ Class Files, ~Vol.~14,  No.~8,  August~2015}%
{Shell \MakeLowercase{\textit{et al.}}: Bare Demo of IEEEtran.cls for IEEE Transactions on Magnetics Journals}
%



\IEEEtitleabstractindextext{%
\begin{abstract}

In the real world,  where information is abundant and diverse across different modalities,  understanding and utilizing various data types to improve retrieval systems is a key focus of research. Multimodal composite retrieval integrates diverse modalities such as text,  image and audio,  etc. to provide more accurate,  personalized,  and contextually relevant results. To facilitate a deeper understanding of this promising direction,  this survey explores multimodal composite editing and retrieval in depth,  covering image-text composite editing,  image-text composite retrieval,  and other multimodal composite retrieval. In this survey,  we systematically organize the application scenarios,  methods,  benchmarks,  experiments,  and future directions.  Multimodal learning is a hot topic in large model era, and have also witnessed some surveys in multimodal learning and vision-language models with transformers published in the PAMI journal. To the best of our knowledge,  this survey is the first comprehensive review of the literature on multimodal composite retrieval, which is a timely complement of multimodal fusion to existing reviews. To help readers' quickly track this field, we build the project page for this survey, which can be found at \href{https://github.com/fuxianghuang1/Multimodal-Composite-Editing-and-Retrieval}{https://github.com/fuxianghuang1/Multimodal-Composite-Editing-and-Retrieval}.

\end{abstract}

\begin{IEEEkeywords}
Multimodal composite retrieval,  Multimodal fusion,  Image retrieval,  Image editing.
\end{IEEEkeywords}}

\maketitle

\IEEEdisplaynontitleabstractindextext

%
\IEEEpeerreviewmaketitle

%
%
%
%

\section{Introduction}\label{introduction}      
    
\IEEEPARstart{I}{n} today's digital landscape,  information is conveyed through various channels such as text,  images,  audio and radar,  etc. resulting in a significant increase in data volume and complexity. As data expands exponentially,  the challenge of processing and integrating diverse information becomes critical. Efficient retrieval of personalized and relevant information is increasingly challenging. 

Traditional unimodal retrieval methods  \cite{huang2020probability,  dubey2021decade,  huang2021domain,  fu2022cross,  zhang2024open,  huang2023coarse,  zhou2023stochastic,  zhang2019optimal,  zhang2020deep,  zhen2019deep,  chun2021probabilistic} depend on a single modality,  such as images or text,  as queries. However,  these approaches often struggle to fully capture the complexities and subtleties of real-world information-seeking scenarios. This limitation has led to the emergence of multimodal composite image retrieval  \cite{vo2018TIRG,  shin2021RTIC,  baldrati2022combiner,  Chen2020VAL,  huang2022-GA-data-augmentation,  kim2021-DCNet,  Leveraging},  a promising framework that transcends the boundaries of individual modalities. By utilizing the complementary strengths of various data types,  multimodal composite retrieval systems enhance the comprehension of user queries and contexts,  resulting in improved retrieval performance and user satisfaction.

\begin{figure*}[t]
    \centering
    \includegraphics[width=1\linewidth]{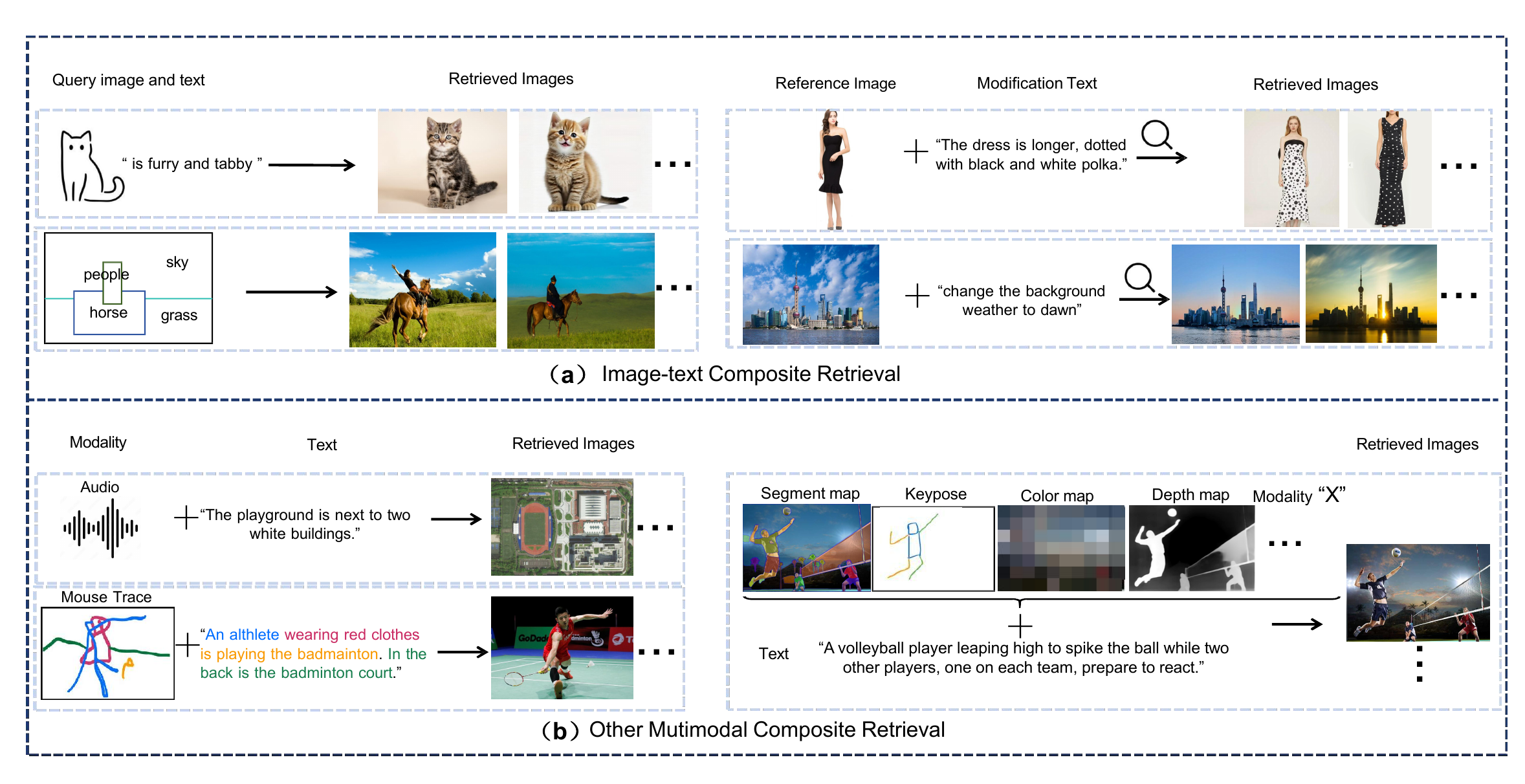}
    \caption{The examples of multimodal composite image retrieval (MCIR) task.}
    \label{examples}
\end{figure*}

As illustrated in Fig. \ref{examples},  multimodal composite retrieval involves the intricate process of merging and analyzing diverse data forms (e.g.,  text,  images,  audio) to retrieve information. This methodology is invaluable across multiple real-world contexts,  including multimedia content \cite{DocStruct4M},  social media platforms,  and e-commerce  \cite{wu2020fashioniq,  han2017fashion200k,  Goenka_2022_FashionVLP,  pang2022MCR,  2019-product-search}. Furthermore,  its applications refer to specialized fields such as medical image retrieval  \cite{2020-DGMFE,  2013-NovaMedSearch,  cao2014medical},  document retrieval  \cite{2013-document-retrieval,  DocStruct4M},  and news retrieval  \cite{tahmasebzadeh2020-news-retrieval}. By employing diverse multimodal queries,  these techniques yield flexible and accurate results,  thereby enhancing user experience and facilitating informed decision-making. Consequently,  multimodal composite retrieval possesses significant potential and research value in information science,  artificial intelligence,  and interdisciplinary applications.

Most existing multimodal composite retrieval methods  \cite{vo2018TIRG,  Chen2020VAL,  chen2020JVSM,  anwaar2021composeAE,  hosseinzadeh2020-locally-LBF,  kim2021-DCNet,  Lee2021CoSMo,  huang2022-GA-data-augmentation,  huang2023-DWC,  liu2024-BLIP4CIR2,  baldrati2022combiner} primarily focus on integrating images and text to achieve desired outcomes. Early methods employed Convolutional Neural Networks (CNNs) for image encoding and Long Short-Term Memory (LSTM) networks  \cite{kiros2014-LSTM-unifying} for text encoding. With the rise of powerful transformers,  such as Vision Transformer (ViT)  \cite{tumanyan2022splicing-ViT},  Swin Transformer (Swin)  \cite{liu2021swin-transformer},  and BERT  \cite{devlin2019BERT},  numerous transformer-based multimodal composite retrieval methods  \cite{tian2022AACL,  ComqueryFormer} have been proposed to enhance image retrieval performance. Additionally,  Vision-Language Pre-training (VLP)  \cite{radford2021CLIP,  li2022BLIP,  li2023-BLIP2,  ALIGN} has transformed tasks related to image understanding and retrieval by bridging the semantic gap between textual descriptions and visual content. Various VLP-based multimodal composite image retrieval methods  \cite{huang2023-DWC,  liu2024-BLIP4CIR2,  baldrati2022combiner} have shown promising results. Furthermore,  image-text composite editing methods  \cite{dong2017-SISGAN,  Zhang_2021-TIM-GAN,  li2020lightweightGAN,  liu2020-DWC-GAN,  cheng2020-SeqAttnGAN,  LS-GAN,  li2020-ManiGAN,  Segmentation-Aware-GAN,  patashnik2021-StyleCLIP} allow users to modify images or generate new content directly through natural language instructions,  achieving precise retrieval that aligns with user intentions. The exploration of additional modalities,  such as audio \cite{2003-SIMC} and motion \cite{yin2024-LAVIMO},  is also gaining momentum.

\textbf{Motivation.} Despite extensive research on multimodal composite retrieval models,  new challenges continue to emerge and remain unresolved. There is a pressing need for comprehensive and systematic analysis in this rapidly evolving field. This survey aims to facilitate a deeper understanding of multimodal composite editing and retrieval by systematically organizing application scenarios,  methods,  benchmarks,  experiments,  and future directions. We review and categorize over 130 advanced methods in multimodal composite retrieval,  providing a solid foundation for further research.
     
\textbf{Literature Collection Strategy.} To ensure a thorough overview of multimodal composite retrieval,  we adopted a systematic search strategy that covers a wide range of relevant literature. Our focus includes studies on innovative methodologies,  applications,  and advancements in multimodal retrieval systems. We selected keywords such as ``multimodal composite retrieval, '' ``multimodal learning, '' ``image retrieval, '' ``image editing, '' and ``feature fusion'' to encompass various facets of this field. These terms reflect foundational concepts,  specific techniques,  and emerging trends commonly found in multimodal research. We conducted searches across prominent academic databases,  including Google Scholar,  DBLP,  ArXiv,  ACM and IEEE Xplore. This exploration yielded diverse sources,  including journal articles,  conference proceedings,  and preprints. To refine our selection,  we excluded studies primarily focused on unimodal approaches or unrelated modalities and manually reviewed the remaining literature for relevance and quality. The final selection process involved evaluating each paper based on its contributions and impact,  enabling us to curate key studies for in-depth analysis. By applying these criteria,  we aim to provide a comprehensive perspective on the current landscape and future directions of multimodal composite retrieval.

\begin{figure*}[t]
    \centering
    \includegraphics[width=1.0\linewidth]{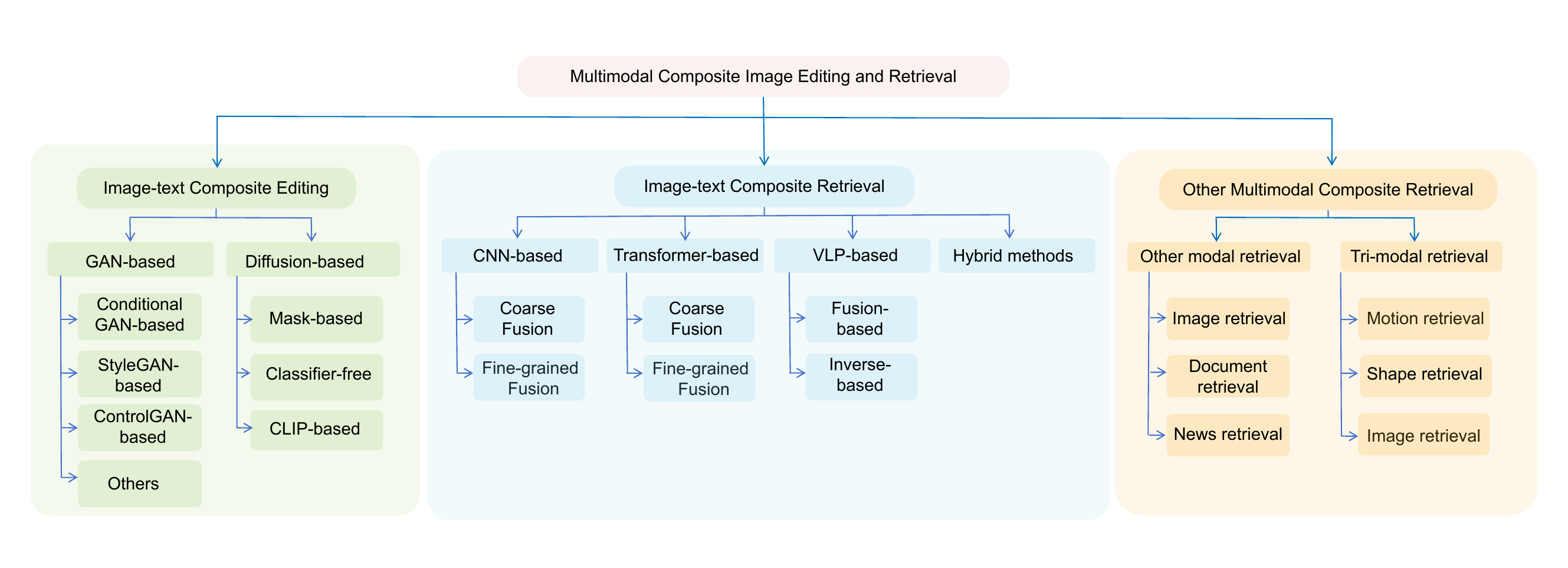}
    \caption{A new taxonomy of multimodal composite editing and retrieval approaches, from three orthogonal aspects in this survey.}
    \label{taxonomy}
\end{figure*}

\textbf{Taxonomy.}  To clarify our discussion on related work in multimodal composite editing and retrieval,  we taxonomize them into three categories through application scenarios in this survey,  i.e.,  1) image-text composite editing,  2) image-text composite retrieval and 3) other multimodal composite retrieval,  as illustrated in Fig. \ref{taxonomy}.  Specifically,   image-text composite editing involves modifying images or creating entirely new content using natural language instructions,  which allows users to clearly and intuitively convey their intentions. Image-text composite retrieval involves searching for personalized results by inputting both text and image information,  which enhances the search experience by enabling users to locate relevant images based on textual descriptions or generate descriptive text from images. Other multimodal composite retrieval tasks feed various combinations of different modalities,  such as audio,  motion,  and other modalities as inputs,  which provides richer and more flexible context-aware retrieval experiences.

\textbf{Contribution.} In summary,  our contributions are as follows: 
\begin{itemize} 
\item To the best of our knowledge,  this paper is the first comprehensive review of multimodal composite retrieval,  aiming to offer a timely overview and valuable insights for future research in this rapidly evolving field. 
\item We systematically organize research achievements,  technical approaches,  benchmarks,  and experiments to enhance understanding of the topic and provide extensive coverage of existing studies with a multi-level taxonomy to cater to the diverse needs of readers. 
\item We address the challenges and open questions in multimodal composite retrieval,  identifying emerging trends and proposing actionable future research directions that can stimulate innovation in this area. \end{itemize}

\textbf{Paper organization.} The rest of the paper is organized as follows. 
In Section \ref{preliminary},  we introduce foundational concepts and applications related to multimodal composite retrieval and establish the context for the methodologies discussed. Section \ref{methods} delves into various methods used in this field,  and categorizes them in terms of their underlying principles and analyzes their strengths and weaknesses. In Section \ref{benchmark},  we present an overview of benchmarks and experimental setups used to evaluate these methods,  along with the results from recent studies. The discussion in Section \ref{discussion} reflects the current state of multimodal composite retrieval,  highlights challenges and proposes future research directions. Finally,  Section \ref{conclusion} concludes the key findings and the significance of this area for future studies.

\section{Preliminary Concepts and Application} \label{preliminary}

\subsection{Preliminary Concepts}
As illustrated in Fig. \ref{examples},  multimodal composite retrieval aims to improve information retrieval flexibility and the overall user experience by integrating text,  image,  and other data forms. The core strength of this technology lies in leveraging the complementary advantages of different data modalities to meet diverse user demands and cognitive preferences. 

\textbf{Image-Text Composite Retrieval.}
Fig. \ref{examples} (a) depicts the image-text composite retrieval process,  which integrates image and text modalities as input to retrieve target images. Specifically,  the input consists of a reference image paired with descriptive text,  which provides guidance for retrieving the target image. The reference image encapsulates complex abstract information,  effectively capturing details such as color,  texture,  and spatial relationships. 
Conversely,  language enables detailed and explicit descriptions,  allowing for the articulation of specific attributes,  relationships,  and context that may not be immediately apparent in an image. By synergistically combining these complementary modalities,  
the system could construct a more comprehensive representation of the target object. 

\textbf{Other Multimodal Composite Retrieval.}
Fig. \ref{examples} (b) illustrates a broader spectrum of multimodal composite retrieval,  extending beyond image and text to include additional modalities such as audio,  mouse traces,  segmentation maps,  key poses,  color maps,  and depth maps.  This integration provides a more nuanced understanding of the user’s search intention,  significantly enhancing the precision and relevance of the retrieved results. 
By leveraging the comprehensive data offered from different modalities,  the system can be well-equipped to accurately identify and retrieve target information.

\subsection{Application Scenarios}
The applications of multimodal composite retrieval are extensive and encompass multiple industries and domains. Several potential applications for multimodal composite retrieval technology are as follows.

\textbf{Fashion and E-commerce.} The integration of text and image modalities shows considerable potential in the fashion industry  \cite{han2017fashion200k,  yuan2021multiturn}. This approach accommodates various cognitive preferences and individual requirements,  allowing users to search for items such as clothing based on specific characteristics like color,  pattern,  and style.

\textbf{Medical Diagnostics.} In the healthcare sector,  multimodal retrieval systems  \cite{cao2014medical} can aid clinicians in locating pertinent images or case studies by merging specific textual descriptions with patient scans,  thus facilitating more accurate diagnoses and informed treatment planning.

\textbf{Smart Cities and Traffic Management.} City management systems can integrate video surveillance,  captured images,  and remote sensing data to swiftly retrieve relevant pictures or videos through text queries (e.g.,  \textit{a person wearing a red shirt} or \textit{the most recent traffic accident}). This system can also amalgamate sensor data to provide a comprehensive situational analysis,  applicable to traffic management,  target searches,  and emergency response.

\textbf{Smart Homes and Personalized Services.} In a smart home setting,  users can articulate their desired atmosphere through voice commands (e.g.,  \textit{romantic dinner setting}),  allowing the system to retrieve and play music or videos that align with the requested ambiance.

\textbf{Content Creation.} Designers can describe a design concept,  prompting the system to automatically retrieve and combine related sketches  \cite{dataset-sketchy},  color schemes  \cite{mou2023-T2I-Adapter},  and audio samples  \cite{2022-MMFR} to generate a series of creative proposals. For instance,  by providing a simple sketch and a text description such as \textit{modern office space},  the system can offer immediate feedback.

\textbf{Intelligent Legal Consultation and Document Retrieval.} Users can inquire about legal issues through language,  prompting the system to automatically retrieve relevant legal texts,  case images,  and documents to generate professional legal advice. For complex cases,  the system can swiftly compile related case laws and legal interpretations based on multimodal inputs.


\textbf{News Scenarios.} Text search functionality can enable users to quickly access real-time trending news,  review historical events,  compile topical reports,  and potentially achieve personalized news recommendations  \cite{tahmasebzadeh2020-news-retrieval}. 

In summary,  multimodal composite retrieval is a highly versatile technology with significant potential for broad applications. It not only enhances the accuracy of information retrieval and user experience but also provides crucial support for personalized and context-aware applications. As technology continues to evolve,  multimodal composite retrieval is playing an increasingly important role across various fields.

\section{Methods}\label{methods}

\subsection{Image-Text Composite Editing}\label{sec3}
Image-Text Composite Editing (ITCE)  manipulates specific elements within an image based on a given text prompt,  which is closely-related with image-text composite retrieval. This is known as text-conditioned image generation,  selectively modifying parts of the image related to textual input while leaving unrelated areas intact.  Due to its versatility and potential for iterative enhancement,  ITCE has wide-ranging applications across various fields.  Two major categories for ITCE include generative adversarial networks (GANs) and diffusion models,  as shown in Table \ref{tableITCE}. 

\begin{table}[t]
\centering
\caption{Methods and architectures for image-text composite editing.}
\huge
\resizebox{\linewidth}{!}{
\begin{tabular}{lccccc}
\hline
\textbf{Methods}
& \textbf{Network}
& \textbf{Image encoder}
& \textbf{Text encoder}
& \textbf{Year}
\\
\hline
SISGAN \cite{dong2017-SISGAN}
& c-GAN
& VGG-16
& LSTM \cite{kiros2014-LSTM-unifying}
& 2017  
\\
GEI \cite{wang2018-GEI}
& c-GAN
& VGG-16
& GRU,  Graph RNN
& 2018  
\\
BRL \cite{Mao_2019-BRL}
& c-GAN
& VGG-16
& LSTM \cite{kiros2014-LSTM-unifying}
& 2019  
\\
SeqAttnGAN \cite{cheng2020-SeqAttnGAN}
& c-GAN
& ResNet-101
& Bi-LSTM
&2020
\\
TIM-GAN \cite{Zhang_2021-TIM-GAN}
& c-GAN
&  -
& BERT
& 2021 
\\
GeNeVA-GAN \cite{elnouby2019-GeNeVA-GAN}
& GAN
& CNN
& GRU
& 2019 
\\
IR-GAN \cite{IR-GAN}
& GAN
& CNN
& GRU
& 2020   
\\
TAGAN \cite{nam2018TA-GAN}
& GAN
& VGG-16
& GRU
& 2018 
\\
lightweightGAN \cite{li2020lightweightGAN}
& GAN
& Inception-v3,  VGG-16
& LSTM
& 2020
\\
FocusGAN \cite{FocusGAN}
& GAN
& Inception-v3,  VGG-16
& RNN
& 2021 
\\
DWC-GAN \cite{liu2020-DWC-GAN}
& GAN
& ResNet-50
& LSTM
& 2020 
\\
CAFE-GAN \cite{Kwak_2020-CAFE-GAN}
& GAN
& -
& -
& 2020 
\\
LS-GAN \cite{LS-GAN}
& GAN
& CNN
& GRU
& 2022 
\\
SegmentationGAN \cite{segmentationGAN}
& GAN
& VGG-16,  ResNet-50
& Transformer
& 2023
\\
StyleCLIP \cite{patashnik2021-StyleCLIP}
& StyleGAN
& -
& -
& 2021 
\\
FFCLIP \cite{zhu2022-FFCLIP}
& StyleGAN 
& e4e \cite{tov2021-GAN-inversion}
& CLIP
& 2022
\\
CLIP2StyleGAN \cite{abdal2022clip2stylegan}
& StyleGAN
& -
& CLIP
& 2022
\\
HairCLIP \cite{wei2022-HairCLIP}
& StyleGAN
& CLIP
& -
& 2022 
\\
DeltaEdit \cite{lyu2023-DeltaSpace}
& StyleGAN
& CLIP
& CLIP
& 2023
\\
FEAT \cite{hou2022-FEAT}
& StyleGAN2
& CLIP
& -
& 2022
\\
TIERA \cite{region-based-attention-TIERA}
& StyleGAN2
& CLIP
& CLIP
& 2023
\\
StyleMC \cite{kocasari2021StyleMC}
& StyleGAN2
& CLIP
& CLIP
& 2022
\\
Paint by word \cite{andonian2023paint-by-word}
& StyleGAN2,  BigGAN
& CLIP
& CLIP
& 2021 
\\
VQGAN-CLIP \cite{crowson2022VQGAN-CLIP}
& VQGAN
& CLIP
& -
& 2022
\\
Segment-aware-GAN \cite{Segmentation-Aware-GAN}
& ManiGAN
& Inception-v3,  VGG-16
& LSTM
& 2021 
\\
ManiGAN \cite{li2020-ManiGAN}
& ControlGAN \cite{li2019controlGAN}
& Inception-v3,  VGG-16
& RNN
& 2020
\\
IIM \cite{shinagawa2018-IIM}
& DCGAN \cite{radford2016-DCGAN}
& -
& LSTM
& 2018 
\\
RAM \cite{chen2018-RAM}
& -
& VGG-16
& LSTM
& 2018 
\\
Open-Edit \cite{liu2021-OpenEdit}
& -
& ResNet
& LSTM
& 2020 
\\
DE-net \cite{tao2022-DENet}
& -
& -
& LSTM
& 2022 
\\
\hline
\end{tabular}
}
\label{tableITCE}
\end{table}

\subsubsection{GAN-based Methods}
  
\textbf{\textit{Conditional GAN-based Methods.}} 
In the category of GAN-based methods,  conditional GANs (cGANs)  \cite{mirza2014-conditionalGAN} utilize additional information (e.g.,  text guidance) as conditioned inputs to generate specific images. We categorize cGAN-based methods into two categories: single-turn generation approaches  \cite{dong2017-SISGAN, Mao_2019-BRL, wang2018-GEI, Zhang_2021-TIM-GAN, FocusGAN, li2020lightweightGAN, liu2020-DWC-GAN, nam2018TA-GAN} and multi-turn generation approaches  \cite{cheng2020-SeqAttnGAN, IR-GAN, chen2018-RAM, LS-GAN}.

\textbf{Single-Turn Generation}.
Most existing image-text composite editing tasks are static,  single-turn generation  \cite{dong2017-SISGAN, Mao_2019-BRL, wang2018-GEI, Zhang_2021-TIM-GAN, FocusGAN, li2020lightweightGAN, liu2020-DWC-GAN, nam2018TA-GAN}. Among them,   \cite{dong2017-SISGAN, wang2018-GEI, Mao_2019-BRL} focus on enhancing the generator \(G\) component. To be specific,  SISGAN  \cite{dong2017-SISGAN} utilizes an encoder-decoder architecture and a residual transformation unit within the generator,  where the encoder and the transformation unit encode combined features of the image and text,  based on which the decoder synthesizes images.
GEI  \cite{wang2018-GEI} investigates three distinct generator architectures,  including a bucket-based model with individual encoder-decoder structure,  grouping similar image transformations,  an end-to-end model featuring a single encoder-decoder for images and a recurrent neural network (RNN) for text,  and a filter-bank model that specifies transformations using trained convolutional filters.
BRL  \cite{Mao_2019-BRL} employs a Bilinear Residual Layer as a conditional layer within the generator to improve representation learning. This network consists of an encoding module,  a fusion module that integrates the semantics of multiple modalities,  a decoding module,  and a discriminator that acts as a classifier to determine whether the generated image aligns with the text description. 
 \cite{nam2018TA-GAN, li2020lightweightGAN, FocusGAN} focus on discriminator \(D\) enhancement. TAGAN  \cite{nam2018TA-GAN} employs a text-adaptive discriminator that evaluates the alignment of text descriptions with images at the word level,  which enables fine-grained modifications that precisely target text-related areas while preserving unrelated regions. LightweightGAN  \cite{li2020lightweightGAN} adopts a lightweight structure with fewer parameters,  including a novel word-level discriminator. It utilizes two distinct image encoders to capture both coarse and detailed information. FocusGAN  \cite{FocusGAN} incorporates a Subject-Focusing Attention (SFA) module to prioritize text-related subjects,  a word-level discriminator to discern fine-grained semantic changes and employs a Background-Keeping Cyclic Loss to maintain background consistency.  
 \cite{liu2020-DWC-GAN} focus on improvement on both the generator \(G\) and the disctriminator \(D\),  operating under the premise that each image can be decomposed into a domain-invariant content space and a domain-specific attribute space  \cite{huang2018-disentangle-assumption2, gonzalezgarcia2018-disentangle-assumption1, liu2020-disentangle-assumption3}. It models high-dimensional content features to improve generation performance. 
Specifically,  TIM-GAN \cite{Zhang_2021-TIM-GAN} models the text as neural operators to modify the input image in the feature space. It synthesize the edited from the image feature modified by the text operator on a predicted spatial attention mask.

\textbf{Multi-Turn Generation}. Multi-turn generation approaches  \cite{cheng2020-SeqAttnGAN, IR-GAN, chen2018-RAM, LS-GAN} feature iterative modifications through a series of instructions,  carried out in multiple steps. SeqAttnGAN  \cite{cheng2020-SeqAttnGAN} employs a neural state tracker to encode the previous image and corresponding text at each step in the sequence,  utilizing a sequential attention mechanism. 
RAM  \cite{chen2018-RAM} utilizes recurrent attentive models to integrate image and language features. It introduces a termination gate for each image region,  which dynamically determines whether to continue extracting information from the textual description after each inference step.
Long and Short-term Consistency Reasoning Generative Adversarial Network (LS-GAN)  \cite{LS-GAN} features a Context-aware Phrase Encoder (CPE) and a Long-Short term Consistency Reasoning (LSCR) module,  capturing long-term visual changes and aligning newly added visual elements with linguistic instructions. 
IR-GAN  \cite{IR-GAN} includes a reasoning discriminator to evaluate the consistency between existing visual elements,  visual increments,  and corresponding instructions. 
    
\textbf{\textit{StyleGAN-based methods.}}
StyleGANs  \cite{karras2019-StyleGAN,  karras2020-StyleGAN2} generate high-quality images by operating within well-disentangled latent spaces,  which is renowned for its capability to produce high-fidelity images. 
Many approaches  \cite{patashnik2021-StyleCLIP,  wei2022-HairCLIP,  kocasari2021StyleMC,  zhu2022-FFCLIP,  hou2022-FEAT,  region-based-attention-TIERA,  lyu2023-DeltaSpace} leverage StyleGAN's latent space  \cite{karras2020-StyleGAN2} to effectively disentangle and manipulate both coarse and fine visual features. For example,   \cite{region-based-attention-TIERA} embeds textual information into the latent space and enhances editing performance by modifying latent codes and searching for manipulation directions,  and interpolates latent vectors within pre-trained GAN models  \cite{wu2020stylespace-editing,  patashnik2021-StyleCLIP}. Traditional methods often require large amounts of labeled data to identify meaningful directions in GAN latent space,  which ecessitates considerable human effort. Leveraging CLIP's powerful image-text representation capabilities can help relieve this problem.
 \cite{patashnik2021-StyleCLIP, wei2022-HairCLIP, andonian2023paint-by-word, segmentationGAN, kocasari2021StyleMC, zhu2022-FFCLIP, lyu2023-DeltaSpace, region-based-attention-TIERA} combine StyleGAN's image generation ability and CLIP's universal image-text representation ability to identify editing directions. These StyleGAN-based methods can be classified into two categories: ``without mask"  \cite{patashnik2021-StyleCLIP,  kocasari2021StyleMC,  zhu2022-FFCLIP,  abdal2022clip2stylegan,  lyu2023-DeltaSpace,  xia2021TediGAN} and ``with mask"  \cite{andonian2023paint-by-word,  hou2022-FEAT,  wei2022-HairCLIP,  region-based-attention-TIERA, xu2022-PPE, segmentationGAN} according to whether the methods use masks to guide the generative model.

As to those methods without additional masks,  StyleCLIP  \cite{patashnik2021-StyleCLIP} introduces three strategies for image-text composite editing,  e.g. latent optimization,  latent mapper,  and global directions. To be specific,  latent optimization adjusts the image's latent code by minimizing the loss in CLIP space to semantically align with the given text. Latent mapper involves training a network to predict a manipulation step in latent space,  which varies depending on the starting position. And Global direction converts a text prompt into a universal mapping direction in latent space,  enabling fine-grained and disentangled visual edits. 
TediGAN  \cite{xia2021TediGAN} encodes both image and text into the latent space to perform style mixing. 
StyleMC  \cite{kocasari2021StyleMC} fine-tunes on a per-prompt basis,  discovering stable global directions by combining CLIP loss with identity loss.
Traditionally,  latent mappings between these two spaces have been manually crafted,  which limits each manipulation model to a specific text prompt. To overcome this limitation,  FFCLIP  \cite{zhu2022-FFCLIP} introduces Free-Form CLIP (FFCLIP),  a method that creates an automatic latent mapping with a cross-attention mechanism that involves semantic alignment and injection,  enabling a single manipulation model to handle free-form text prompts. 
DeltaEdit  \cite{lyu2023-DeltaSpace} incorporates CLIP DeltaSpace,  which semantically aligns the visual feature differences between two images with the textual feature differences in their corresponding descriptions. 
CLIP2StyleGAN  \cite{abdal2022clip2stylegan} connects the pretrained latent spaces of StyleGAN and CLIP to automatically derive semantically labeled editing directions within StyleGAN. It achieves this by leveraging the CLIP image space to identify potential edit directions,  using the CLIP text encoder to disentangle and label these directions,  and then mapping the labeled,  disentangled directions back to the StyleGAN latent space to enable various unsupervised semantic modifications.
    
 \cite{andonian2023paint-by-word,  hou2022-FEAT,  wei2022-HairCLIP,  region-based-attention-TIERA, xu2022-PPE, segmentationGAN} use masks to accomplish manipulation. HairCLIP  \cite{wei2022-HairCLIP} first obtains the latent code of the input image using the StyleGAN inversion method “e4e”  \cite{tov2021-GAN-inversion},  then predicts the latent code changes and editing conditions using a mapper network,  and ultimately feeds the modified latent code back into the pre-trained StyleGAN to generate images.
Paint by Word  \cite{andonian2023paint-by-word} utilizes CLIP to provide feedback on the generated images,  performing manipulations within a user-specified region based on a given mask. 
TIERA  \cite{region-based-attention-TIERA} utilizes a region-based spatial attention mechanism to identify the editing area accurately. It begins by encoding the text input using CLIP,  then employs a mapping module to adjust the original image's style codes based on the text embedding. 
SegmentationGAN  \cite{segmentationGAN} employs the referring image segmentation to determine text-relevant and irrelevant regions,  using CLIP as a loss function to ensure consistency between modified and unchanged areas.
Unlike earlier methods that heavily rely on disentangling various attributes in the latent space,  FEAT  \cite{hou2022-FEAT} employs learned attention masks to concentrate on edited areas and limits modifications to specific spatial regions and PPE  \cite{xu2022-PPE} predicts potentially entangled attributes corresponding to a specified text command first,  and then introduces an disentanglement loss. 

\textbf{\textit{ControlGAN-based methods.}}
ControlGAN  \cite{li2019controlGAN} enables the synthesis of high-quality images while allowing control over specific aspects of the generation process based on natural language descriptions.
ManiGAN  \cite{li2020-ManiGAN} builds on the multi-stage architecture of ControlGAN by introducing a multi-tiered framework that includes a text-image affine combination module (ACM) and a detail correction module (DCM). 
Segmentation-aware GAN  \cite{Segmentation-Aware-GAN} incorporates an image segmentation network into the generative adversarial framework,  similar to ManiGAN  \cite{li2020-ManiGAN}. The segmentation encoder is based on the pre-trained Deeplabv3  \cite{chen2018-deeplabv3} and detects the foreground and background of the input image,  improving the model's ability to generate contextually accurate and visually coherent images.

\textbf{\textit{Other GAN-based methods.}}
Creating and editing images from open-domain text prompts has been challenging,  often requiring expensive and specially designed models. VQGAN-CLIP  \cite{crowson2022VQGAN-CLIP} adopts an innovative approach by using CLIP to guide VQGAN,  adjusting the similarity between candidate generations and the guiding text. OpenEdit  \cite{liu2021-OpenEdit} is the first to explore open-domain image editing with open-vocabulary instructions. 
DE-Net  \cite{tao2022-DENet} dynamically assembles various editing modules to accommodate different editing needs. 
CAFE-GAN  \cite{Kwak_2020-CAFE-GAN} focuses on editing facial regions relevant to target attributes by identifying specific areas with target and complementary attributes. 
IIM  \cite{shinagawa2018-IIM} constructs a neural network that operates on image vectors within the latent space,  transforming the source vector into the target vector using an instruction vector.

\subsubsection{Diffusion-based Methods}
We categorize diffusion-based methods according to the guidance mechanisms,  i.e. mask-based methods  \cite{couairon2022-DiffEdit, wang2023-InstructEdit, cao2023masactrl, choi2023-CustomEdit, zhang2023IIR, nichol2022-Glide, Avrahami_2022-Blended-Diffusion, park2024-shape-guided, ravi2023PRedItOR},  classifier-free methods  \cite{dong2023-PTI, zhang2022SINE, kawar2023-Imagic, valevski2023-Unitune, hertz2022prompttoprompt, wang2023-MDP, brooks2023-InstructPix2Pix, tumanyan2022-DF},  and CLIP-based methods  \cite{kim2022-DiffusionCLIP,  kwon2023-DiffuseIT}.

\textbf{\textit{Mask-based methods. }}
Mask-based methods utilize masks to localize specific areas for modification. For instance,  Blended Diffusion  \cite{Avrahami_2022-Blended-Diffusion} employs masks for image-text composite editing in either pixel or latent space. Subsequent work  \cite{couairon2022-DiffEdit, park2024-shape-guided, liu2023-DINO, wang2023-InstructEdit} has automatic mask generation using cross-attention maps,  replacing manual masks with automatic ones. These methods can be further divided into manual mask  \cite{nichol2022-Glide, Avrahami_2022-Blended-Diffusion},  automatic mask  \cite{couairon2022-DiffEdit, wang2023-InstructEdit, cao2023masactrl, choi2023-CustomEdit, zhang2023IIR},  and optional mask  \cite{park2024-shape-guided, ravi2023PRedItOR} approaches.

\textbf{Manual mask}. 
Glide  \cite{nichol2022-Glide} compares CLIP guidance and classifier-free guidance,  finding that the latter is preferred for its ability to leverage internal knowledge for guidance,  thereby simplifying conditioning processes that classifiers often struggle with. Blended  Diffusion  \cite{Avrahami_2022-Blended-Diffusion} combines CLIP guidance with a denoising diffusion probabilistic model to seamlessly blend edited and untouched image regions by introducing noise at various levels. 
    
\textbf{Automatic mask}.
InstructEdit  \cite{wang2023-InstructEdit} employs automatic masking for precise edits by using ChatGPT and BLIP2  \cite{li2023-BLIP2} to convert text instructions into a segmentation prompt,  input caption,  and edited caption,  using Grounded Segment Anything,  which combines Segment Anything  \cite{kirillov2023segmentanything} and Grounded DINO  \cite{liu2023-DINO} to generate masks,  and using Stable Diffusion  \cite{rombach2022-stablediffusion} to finalize the edited image. DiffEdit  \cite{couairon2022-DiffEdit} automatically infers a mask to guide the denoising process in a text-conditioned diffusion model,  minimizing unintended edits. Shape-Guided Diffusion  \cite{park2024-shape-guided} generates object masks from prompts and employs Inside-Outside Attention to constrain attention maps. Custom-Edit  \cite{choi2023-CustomEdit} customizes diffusion models by optimizing language-relevant parameters and applies P2P  \cite{hertz2022prompttoprompt} and Null-text inversion techniques  \cite{mokady2022nulltext} for precise edits. IIR  \cite{zhang2023IIR} introduces an Image Information Removal module to preserve non-text-related content while enhancing text-relevant details.


\textbf{Optional mask}. 
PRedItOR  \cite{ravi2023PRedItOR} employs a Hybrid Diffusion Model (HDM),  which uses CLIP embeddings for more accurate inversions and enables structure-preserving edits without needing additional inputs or optimization. SDEdit  \cite{meng2022-SDEdit} edits images by starting the sampling process from a noisy version of the base image. However,  this approach is less effective for fine detail restoration,  especially when significant pixel-level changes are required.

\textbf{\textit{Classifier-free methods. }}
Classifer-free methods  \cite{dong2023-PTI, zhang2022SINE, kawar2023-Imagic, valevski2023-Unitune, hertz2022prompttoprompt, wang2023-MDP, brooks2023-InstructPix2Pix, tumanyan2022-DF} refer to guiding the generation process by directly adjusting the results from both conditional and unconditional model outputs,  instead of using a pre-trained classifier to steer the diffusion process.
To mitigate overfitting issues when fine-tuning pre-trained diffusion models on a single image,  SINE \cite{zhang2022SINE} introduces a novel model-based guidance approach built on classifier-free guidance,  which distills the knowledge acquired from a model trained on a single image into the pre-trained diffusion model. 
Prompt-to-Prompt \cite{hertz2022prompttoprompt} enhances editing quality by leveraging the visual-semantic data encoded in the intermediate attention matrices of a text-to-image model. However,  this technique relies on attention weights,  limiting its application to images generated by the diffusion model. 
MasaCtrl \cite{cao2023masactrl} enhances text-image consistency by transforming traditional self-attention in diffusion models into mutual self-attention. 
Imagic \cite{kawar2023-Imagic},  a pre-trained text-to-image diffusion model,  begins by optimizing a text embedding to produce images that resemble the input image. 
InstructPix2Pix \cite{brooks2023-InstructPix2Pix} combines the strengths of GPT-3 \cite{brown2020-GPT-3} and Stable Diffusion \cite{rombach2022-stablediffusion} to create an image-editing dataset that captures complementary knowledge from both language and images. 
The success of this training process is highly dependent on the quality of the dataset and the performance of the diffusion model.
Unitune \cite{valevski2023-Unitune} builds on the idea that image-generation models can be adapted for image editing through fine-tuning on a single image. 
PTI  \cite{dong2023-PTI} designs Prompt Tuning Inversion,  an efficient and accurate technique for text-driven image editing. 
Plug-and-Play \cite{tumanyan2022-DF} is a modern model that utilizes attention maps from intermediate layers to transfer features from one image to another.
MDP  \cite{wang2023-MDP} introduces a framework that delineates the design space for appropriate manipulations,  identifying five distinct types: intermediate latent,  conditional embedding,  cross-attention maps,  guidance,  and predicted noise.

\textbf{\textit{CLIP-based methods.}} 
DiffuseIT  \cite{kwon2023-DiffuseIT} presents a diffusion-based unsupervised image translation method leveraging disentangled style and content representations. Inspired by the Splicing ViT  \cite{tumanyan2022splicing-ViT},  DiffuseIT incorporates a loss function that utilizes intermediate keys from the multihead self-attention layers of a pre-trained ViT model to guide the DDPM model's generation process,  thereby ensuring content preservation and enabling semantic alterations. 
DiffusionCLIP  \cite{kim2022-DiffusionCLIP} employs a deterministic DDIM noising process to accurately identify the specific noise required to generate the target image.

\begin{figure*}[t]
    \centering
    \includegraphics[width=1.0\linewidth]{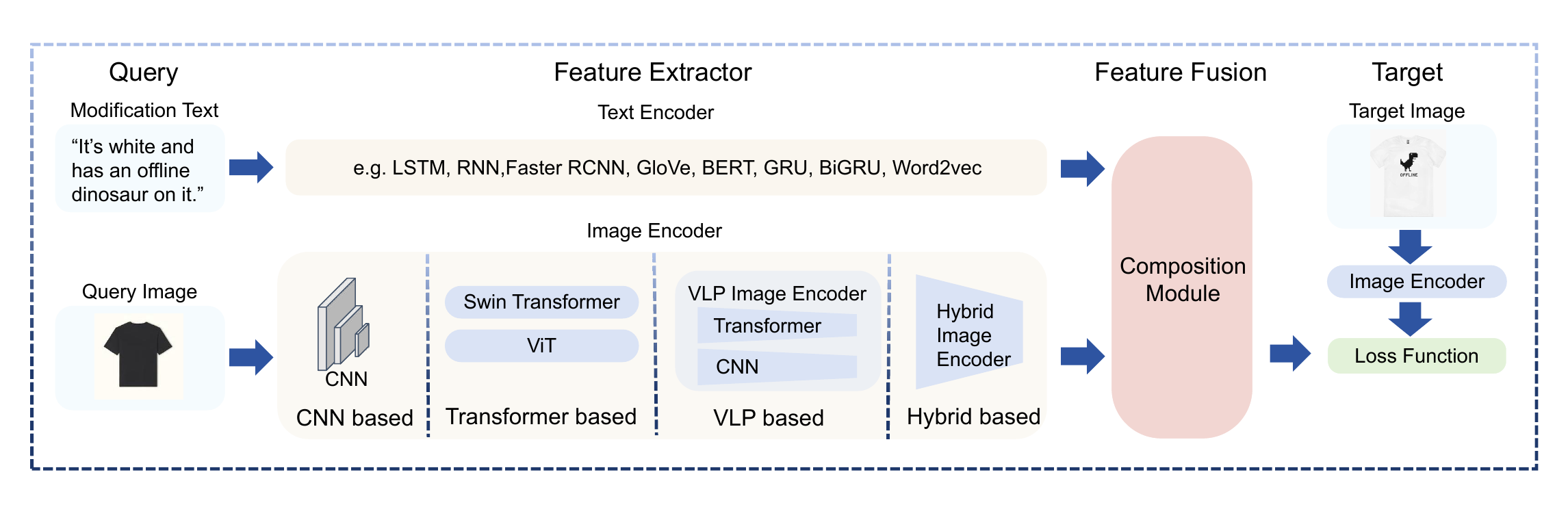}
    \caption{The illustration of the basic technical framework of image-text composite retrieval.}
    \label{illustration}
\end{figure*}

\subsubsection{Summary}


    GANs are renowned for their ability to generate high-fidelity images,  making them a popular choice in image-text composite editing. Key techniques within this approach include disentangling the latent space and optimizing generator parameters to improve cross-modal feature alignment. By leveraging the representational power of CLIP,  GANs can more effectively identify latent directions and measure similarity with text prompts. These capabilities allow for precise and controlled image manipulation based on textual descriptions.
    Diffusion models have recently emerged as a powerful alternative,  excelling in synthesizing high-quality images from noise through iterative denoising. 
    By integrating with various image-text methodologies,  diffusion models have significantly pushed the boundaries of image editing,  particularly in terms of quality and flexibility.
    While GANs are adept at generating high-resolution images with controlled edits and diffusion models offer a more systematic and iterative approach to image generation,  
    especially in complex scenarios,  the key distinction lies in the underlying process and training strategy,  i.e.,  adversarial training of GAN and progressive denoising of diffusion. 

   Some challenges and perspectives are summarized as follows.
\begin{enumerate} 
    \item Consistency Maintenance: Future techniques should focus on maintaining consistency by ensuring that text-irrelevant areas of an image remain unchanged,  while selectively modifying attributes described in the text. This is crucial for preserving overall image coherence during targeted edits. 
    \item Precision Enhancement: Improving precision by enabling the manipulation of specific attributes across multiple objects within an image is essential. This includes refining the granularity of edits,  particularly in complex scenes with multiple objects.
    \item Robustness in Complex Scenarios: Enhancing the robustness of models to execute realistic modifications in open-domain scenarios and complex scenes is another key challenge. As the complexity of scenes increases,  the ability to maintain realism after editing becomes increasingly important.
\end{enumerate}

\subsection{Image-text Composite Retrieval}\label{sec2}

    In the field of image-text composite retrieval,  the objective is to identify a target image by utilizing a reference image alongside textual descriptors that specify differences between reference and target images. The text is used to modify the reference image. Since this task involves aligning reference image with target image by introducing the modification instructions in the text,  the task can also be referred as ``text-guided image retrieval''.
    
    As shown in Fig. \ref{illustration},  a standard framework for composite image retrieval comprises three main components: feature extraction,  image-text composition module,  and alignment. 
    Traditionally,  image representations are derived from the final layer of Convolutional Neural Networks (CNNs) or Vision Transformers (ViTs),  while text encoders typically rely on Recurrent Neural Networks (RNNs),  LSTMs  \cite{graves2012LSTM},  GloVe  \cite{pennington2014-GloVe},  BERT  \cite{devlin2019BERT},  and GRUs  \cite{cho2014GRU}. Recently,  with the advent of large models,  encoders from CLIP  \cite{radford2021CLIP} have become increasingly popular due to their well-aligned text and image representations.
    
    Based on a comprehensive review,  we provide a taxonomy of the image-text composite retrieval methods in terms of the image encoder backbone,  including CNN-based (\textsection \ref{CNN based methods}),  Transformer-based (\textsection \ref{Transformer based methods}),  large model-based (\textsection \ref{Large Model based methods}),  and hybrid methods (\textsection \ref{Hybrid methods}). From a framework perspective,  some methods  \cite{vo2018TIRG, shin2021RTIC, baldrati2022combiner, Chen2020VAL} focus on designing the composition module to enhance performance,  while others  \cite{shin2021RTIC, huang2022-GA-data-augmentation, chen2024uncertainty, kim2021-DCNet, Leveraging} emphasize additional modules to improve performance,  and  \cite{delmas2022ARTEMIS, jandial2021SAC, chen2020JVSM} aim to enhance the overall framework.
    The development of image-text composite retrieval (ITCR) has seen a significant evolution,  transitioning from CNN-based to Transformer-based backbones,  and more recently to large models. This progression has been driven by advances in deep learning within both computer vision and natural language processing. Large-scale pre-trained cross-modal models like CLIP  \cite{radford2021CLIP} and BLIP  \cite{li2022BLIP} have further enhanced ITCR,  leveraging their robust capabilities in multimodal representation. For clarity,  we provide a detaied summary of ITCR methods in Table \ref{ITCR methods}.


\begin{table*}
\centering
\caption{Image-text composite retrieval methods.}
\large
\fontsize{3}{3}\selectfont
\renewcommand{\arraystretch}{0.9}
\resizebox{\linewidth}{!}{
\begin{tabular}{lccccc}
\hline

\textbf{Methods}
& \textbf{Image encoder}
& \textbf{Text encoder}
&\textbf{Composition}
& \textbf{Year}
\\
\hline
DT \cite{han-schlangen-2017-DT}
& GoogleNet
& -
& Late fusion
& 2017
\\
SSIS \cite{SSIS}
& GoogleNet
& Word2Vec
& -
& 2017
\\
ComposeAE \cite{anwaar2021composeAE}
& ResNet-18
& BERT
& Complex projection
& 2021
\\
HCL \cite{HCL}
& ResNet-18
& LSTM
& Hierarchical Composition
& 2021
\\
TIRG  \cite{vo2018TIRG}
& ResNet-18
& LSTM
& Residual Gating
& 2019
\\
JPM \cite{JPM}
& ResNet-18
& LSTM
& Residual Gating,  Transformer 
& 2021
\\
GA \cite{huang2022-GA-data-augmentation}
& ResNet-18
& LSTM/BERT
& Residual Gating
& 2023
\\
DeepStyle \cite{tautkute2019-DeepStyle}
& ResNet-50
& Word2vec \cite{mikolov2013-word2vec}
& Concatenation
& 2019
\\
MAAF \cite{dodds2020MAAF}
& ResNet-50
& LSTM
& Transformer
& 2020
\\
SVSCAL \cite{kilickaya2020-SVSCAL}
& ResNet-50
& -
& -
&  2020
\\
CoSMo \cite{Lee2021CoSMo}
& ResNet-50
& LSTM
& Content \& Style Modules
& 2021
\\
CLVC-Net \cite{wen2021-CLVC-NET}
& ResNet-50
& LSTM
& Attention mechanism
& 2021
\\
RTIC \cite{shin2021RTIC}
& ResNet-50
& LSTM
& Residual Gating
& 2021
\\
DCNet \cite{kim2021-DCNet}
& ResNet-50
& GloVe
& Composition\& Correction
& 2021
\\
Leveraging \cite{Leveraging}
& ResNet-50
& BERT,  GRU
& Content\& Style
& 2021
\\
SAC \cite{jandial2021SAC}
& ResNet-50
& GRU,  BERT
& Residual Gating 
& 2022
\\
MPC \cite{neculai2022-MPC}
& ResNet-50
& GloVe,  BiGRU
& Probabilistic composer
& 2022
\\
EER \cite{EER}
& ResNet-50
& GloVe,  LSTM
& Suppression\& Replenishment
& 2022
\\ 
AMC \cite{AMC}
& ResNet-50
& GloVe,  LSTM
& Adaptive router
& 2023
\\
Uncertainty \cite{chen2024uncertainty}
& ResNet-50
& RoBERTa
& Content\& Style Modules 
& 2024
\\
MCR \cite{pang2022MCR}
& ResNet-50
& LSTM
& Transformer
& 2021
\\
DIIR \cite{guo2018-DIIR}
& ResNet-101
& CNN
& Concatenation
& 2018
\\
LBF \cite{hosseinzadeh2020-locally-LBF}
& ResNet-101
& WordPiece
& Cross-modal attention
&2020
\\
CRR \cite{CRR}
& ResNet-101
& GRU
& Cross-modal attention
& 2022
\\
GSCMR \cite{2022-GSCMR}
& ResNet-101
& Bi-GRU
& Attention mechanism
& 2022
\\
CIRPLANT \cite{liu2021CIRPLANT}
& ResNet-152
& - 
& VLP Multi-layer Transformer
& 2021
\\
ASA \cite{han2017fashion200k}
& Inception-v3
& Word2vec
& -
& 2017
\\
TIS \cite{TIS}
& Inception-v3
& LSTM
& GAN-based
& 2022
\\
JVSM~ \cite{chen2020JVSM}
& MobileNet-v1
& LSTM
& Residual Gating 
& 2020
\\
ARTEMIS \cite{delmas2022ARTEMIS}
& ResNet-18,  ResNet-50
& Glove,  LSTM/BiGRU
& Attention mechanism
& 2022
\\
FashionVLP \cite{Goenka_2022_FashionVLP}
& ResNet-18,  ResNet-50
& BERT
& VinVL \cite{zhang2021vinvl}
& 2022
\\
FashionNTM \cite{pal2023fashionNTM}
& ResNet-18,  ResNet-50
& BERT
& Memory network
& 2022
\\
VAL \cite{Chen2020VAL}
& ResNet-50,  MobileNet
& LSTM
& Transformer
& 2020
\\
DATIR \cite{DATIR}
& ResNet-50,  MobileNet
& LSTM
& Transformer
& 2021
\\
multiturn \cite{yuan2021multiturn}
& ResNet-101, ResNet-152
& GloVe
& Complex projection
& 2021
\\
CurlingNet~ \cite{yu2020Curlingnet}
& ResNet-152,  DenseNet-169
& BiGRU-CNN 
& Context Gating \cite{miech2018-context-gating}
& 2020
\\
Fashion-IQ \cite{wu2020fashioniq}
& EfficientNet-b  \cite{tan2020efficientnet})
& GloVe
& Transformer
& 2021
\\
SceneTrilogy \cite{chowdhury2023-SceneTrilogy}
& VGG-16
& Bi-GRU \cite{cho2014GRU}
& Cross-attention
& 2023 
\\
QSS \cite{rossetto2019-QSS}
& DeepLab
& Word2vec \cite{mikolov2013-word2vec} 
& Aggregation
& 2019 
\\
RBIRR \cite{hinami2017-RBIRR}
& VGG16,  AlexNet
& Fast RCNN
& Joint/Concatenate/Merge
& 2017 
\\

AACL \cite{tian2022AACL}
& Swin 
& DistilBERT
& Additive Attention Composition
& 2021
\\
ProVLA \cite{Hu_2023_ICCV-ProVLA}
& Swin 
& Transformer
& Cross attention
& 2023
\\
CRN \cite{2023-CRN}
& Swin 
& LSTM
& Hierarchical Aggregation Transformer
& 2023
\\
ComqueryFormer \cite{ComqueryFormer}
& Swin 
& BERT
& Cross-modal transformer
& 2023
\\

BLIP4CIR2 \cite{liu2024-BLIP4CIR2}
& BLIP
& BLIP
& Concatenate
& 2024
\\
CASE \cite{levy2023CASE}
& BLIP
& BERT
& Cross-attention
& 2023
\\
BLIP4CIR1 \cite{liu2023BLIP4CIR1}
& BLIP-B
& BLIP
& Concatenate
& 2023
\\
SPRC \cite{bai2023-SPRC-sentencelevel}
& BLIP-2
& BLIP-2
& Inverse based
& 2023
\\
CLIP4CIR2 \cite{CLIP4CIR3}
& CLIP (RN50)
& Transformer
& Combiner function
& 2023
\\
Combiner \cite{baldrati2022combiner}
& CLIP (RN50)
& CLIP
& Combiner function
& 2022
\\
CLIP4CIR1 \cite{baldrati2022-CLIP4CIR,  CLIP4CIR2}
& CLIP (RN50)
& Transformer
& Combiner function
& 2022
\\
CIReVL \cite{karthik2024-CIReVL}
& CLIP
& CLIP
& Inverse based
& 2024
\\
CompoDiff \cite{gu2024compodiff}
& CLIP
& CLIP/T5
& Denoising Transformer
& 2024
\\
TG-CIR \cite{Wen_2023-TG-CIR}
& CLIP-B
& CLIP
& Keep-and-replace
& 2023
\\
TASKformer \cite{sangkloy2022-TASKformer}
& CLIP-B
& CLIP
& Element-wise addition
& 2022 
\\
PALAVRA \cite{cohen2022-PALAVRA}
& CLIP-B
& CLIP-B
& Inverse based
& 2022
\\
PL4CIR \cite{zhao2022-PL4CIR_PLHMQ-twostage}
& CLIP-B
& BERT
& Adaptive Weighting
& 2022
\\
Pic2Word \cite{saito2023pic2word}
& CLIP-L
& CLIP-L
& Inverse based
& 2023
\\
Contexti2w \cite{tang2023contexti2w}
& CLIP-L
& CLIP-L 
& Inverse based
& 2023
\\
KEDs \cite{suo2024KEDs}
& CLIP-L 
& CLIP-L 
& Inverse based
& 2024
\\
Enhancing \cite{Zhu_2024-enhancing}
& CLIP-L
& CLIP-L
& Inverse based
& 2024
\\
SBCIR \cite{koley2024-SBCIR}
& CLIP-L
& CLIP
& Inverse based
& 2024 
\\
DQU-CIR \cite{Wen_2024-DQU-CIR}
& CLIP-H
& CLIP-H
& Inverse based
& 2024
\\
SEARLE \cite{Baldrati2023SEARLE}
& CLIP-B,  CLIP-L
& CLIP-B,  CLIP-L
& Inverse based
& 2023
\\
MagicLens \cite{zhang2024magiclens}
& CLIP-B,  CLIP-L
& CLIP
& Concatenate
& 2024
\\
LLM4MS \cite{barbany2024-LLM4MS}
& CLIP-L,  BLIP2
& T5
& Concatenate
& 2024
\\
Ranking-aware \cite{chen2023ranking-aware}
& CLIP (RN50,  Transformer)
& CLIP
& Concatenate
& 2023
\\
PLI \cite{chen2023-PLI}
& CLIP-B,  CLIP-L,  BLIP-B
& CLIP-B,  CLIP-L,  BLIP-B
& Inverse based
& 2023
\\
LinCIR \cite{gu2024LinCIR}
& CLIP-L,  CLIP-H,  CLIP-G 
& CLIP-L,  CLIP-H,  CLIP-G 
& Inverse based
& 2023
\\

LGLI \cite{huang2023-LGLI}
& ResNet18,  CLIP
& LSTM
& Attention mechanism
& 2023
\\
DWC \cite{huang2023-DWC}
& ResNet50,  CLIP(RN50)
& LSTM
& Editable Modality De-equalizer
&2024
\\
AlRet \cite{xu2024-AlRet}
& ResNet18,  ResNet50,  CLIP
& LSTM,  GloVe
& Attention mechanism
&2024
\\
\hline
 \end{tabular}}
\label{ITCR methods}
\end{table*}

\subsubsection{CNN-based Methods}\label{CNN based methods}

    Convolutional Neural Networks (CNNs)  \cite{krizhevsky2017-CNN-resnet-50} have been pivotal in 
    extracting hierarchical features from images.  \cite{babenko2014CNN-demonstrate} has demonstrated that activations in the upper layers of a CNN serve as sophisticated visual content descriptors of an image. Specifically,  a CNN (e.g.,  AlexNet  \cite{krizhevsky2012Alex},  VGG  \cite{simonyan2015-VGG-16},  ResNet  \cite{krizhevsky2017-CNN-resnet-50},  DenseNet  \cite{huang2017densely},  GoogleNet  \cite{szegedy2015going} and MobileNet  \cite{howard2017mobilenets}) pre-trained on ImageNet  \cite{wang2014-imagenet} can be used to obtain image embedding by employing global pooling in the last CNN layer,  and show remarkable success in various computer vision tasks  \cite{he2015-cnn-based3,  Feng_2021-cnn-based2,  feng2019-cnn-based1,  hu2021-cnn-based4,  wei2019-cnn-based5,  yang2021-cnn-based6}. Consequently,  many CNN-based methods  \cite{vo2018TIRG,  chen2020JVSM,  anwaar2021composeAE,  JPM,  delmas2022ARTEMIS,  huang2022-GA-data-augmentation, hosseinzadeh2020-locally-LBF,  kim2021-DCNet,  Lee2021CoSMo,  chen2024uncertainty,  yu2020Curlingnet,  Leveraging,  wen2021-CLVC-NET,  jandial2021SAC,  AMC,  TIS},  as shown in Table \ref{ITCR methods},  adopt CNN backbones as the image encoder for conducting ITCR task. 
    To achieve more granular feature extraction,  SAC \cite{jandial2021SAC} employs multiple levels to capture both coarse and fine-grained features.  LBF \cite{hosseinzadeh2020-locally-LBF} utilizes Faster R-CNN \cite{ren2016-fasterRCNN} to improve the composition of text and image features. The fusion of these features is commonly categorized into coarse and fine-grained approaches. Coarse fusion,  as proposed in \cite{vo2018TIRG,  chen2020JVSM,  hosseinzadeh2020-locally-LBF,  Lee2021CoSMo,  anwaar2021composeAE,  kim2021-DCNet,  JPM,  delmas2022ARTEMIS,  huang2022-GA-data-augmentation,  chen2024uncertainty,  yu2020Curlingnet},   involve integrating high-level features from each modality into a single,  unified representation,  which enhances retrieval performance by maintaining the overall context. In contrast,  fine-grained fusion,  as proposed in  \cite{wen2021-CLVC-NET,  Leveraging,  jandial2021SAC,  TIS,  AMC},  divide features into separate modules (e.g.,  style and content) and then combine the outputs to form a final representation.

\textbf{\textit{Coarse fusion methods.}}
    Coarse fusion is a technique commonly used in multimodal composite retrieval systems to integrate information. It involves  synthesizing high-level features extracted from each modality into a single,  unified representation. The goal is to capture the critical information from each modality while preserving the overall context,  thus enhancing retrieval performance.

\textbf{Gating Mechanism.} 
    In Text Image Residual Gating (TIRG) \cite{vo2018TIRG},  the task of text-guided image semantic alignment is first proposed by employing a learned gated residual connection and a residual connection,  in order to selectively modifies image features based on the text description while preserving the aspects of the image unrelated to the text. 
    Many subsequent methods \cite{chen2020JVSM, yu2020Curlingnet, JPM} adopt the gating mechanism of TIRG as their composition module. 
    JVSM \cite{chen2020JVSM} jointly learn unified joint visual semantic matching within a visual semantic embedding framework. It seeks to encode the semantic similarity between visual data (i.e.,  input images) and textual data (i.e.,  attribute-like descriptions). 
    CurlingNet \cite{yu2020Curlingnet} designs two networks named Delivery filters and Sweeping filter,  the former transits the reference image in an embedding space and the latter emphasizes query-related components of target images in the embedding space,  which aims to find better ranking of a group of target images.
    DCNet \cite{kim2021-DCNet} introduces the Dual Composition Network,  by taking into account both forward (the composition network) and inverse (the correction network,  which models the difference between the reference and target images in the embedding space and aligns it with the text query embedding) pathways.
    EER  \cite{EER} addresses the composite image retrieval task by methodically modeling two key sub-processes: image semantics erasure and text semantics replenishment. 
    To explore the intrinsic relationships between different modalities,  Yang et al. introduce the Joint Prediction Module (JPM) \cite{JPM}. To alleviate semantic inconsistencies caused by different pre-trained models and distinct latent spaces,  AET \cite{zhang2019AET} views the reference image and the target image as a pair of transformed images and regards the modification text as an implicit transformation. 
    To slove the problem of data scarcity and low generalization,  RTIC \cite{shin2021RTIC} utilizes a graph convolutional neural network (GCN) as a regularizer by facilitating information propagation among adjacent neighbors. 
    Observing that the characteristics of training data significantly influence the training outcomes,  and considering that traditional data often results in overfitting and exhibits a low diversity of training distributions,  data augmentation becomes crucial. Therefore,  Huang et al. \cite{huang2022-GA-data-augmentation} propose a gradient augmentation (GA) model for ITCR,  an implicit data augmentation inspired by adversarial training for resisting perturbations and a rationale that gradient changes can also reflect data changes to some extent. 

\textbf{Attention Mechanism.} 
    LBF \cite{hosseinzadeh2020-locally-LBF} represents the reference image by a set of local entities and establishes relationships between each word of the modification text and these local areas. This approach achieves bidirectional correlation between text and image. It then operates the fusion process by incorporating a cross-modal attention module. 
    JGAN  \cite{JGAN} introduces a unified model that simultaneously manipulates image attributes based on modification text using a jumping graph attention network and an adversarial network to learn text-adaptive representations for queries. 
    ARTEMIS  \cite{delmas2022ARTEMIS} treats the modification text as a distributor of weights across the visual representations of both the reference image and the target image and designs an Explicit Matching module and an Implicit Similarity module. 
    CRR \cite{CRR} introduces a memory-augmented cross-modal attention module for integrating image and text features and two graph reasoning modules to establish intra-modal relationships within the query and the target separately.
    CIRPLANT \cite{liu2021CIRPLANT} is a transformer-based model that utilizes a pre-trained vision-and-language model i.e. Oscar \cite{li2020oscar} and constructs a multi-layer transformer as the composition module to modify visual features. 
    MAAF \cite{dodds2020MAAF} extract vector ``tokens'' that represent elements from each input modality,  and then compile these tokens into a unified sequence  via an attention model.

\textbf{Others.} 
    ComposeAE \cite{anwaar2021composeAE} suggests a model based on auto-encoders 
    and incorporates an explicit rotational symmetry constraint into the optimization process. 
    AMC \cite{AMC} is an Adaptive Multi-expert Collaborative network,  whose routers can dynamically adjust the activation and achieve adaptive fusion of reference image and text embeddings. 
    SceneTrilogy \cite{chowdhury2023-SceneTrilogy} is a unified framework that jointly model sketch,  text,  and photo to seamlessly support several downstream tasks like fine-grained sketch and text based image retrieval. 
    RBIRR \cite{hinami2017-RBIRR} is capable of performing instance retrieval related to multiple objects,  providing the category or attribute of the objects and the position constraints among them,  including spatial location,  size,  and relationship. 
    SSIS \cite{SSIS} adopts a ``first generate,  then retrieve'' paradigm,  training a convolutional neural network to synthesize visual features that capture the spatial-semantic constraints from the user's canvas query.
    EISSIR \cite{furuta2019-EISSIR} is an interactive image retrieval system based on semantic segmentation,  which interprets the segmentation map drawn by user as a binary probability map.

\textbf{\textit{Fine-grained fusion methods.}}
    Given that the guided text can vary from describing concrete attributes,  
    the ability to conduct fine-grained fusion is necessary. It is advantageous to design a framework capable of processing information from multi-layers and conduct fusion using multiple separate modules,  e.g. content and style modules \cite{Leveraging, Lee2021CoSMo} and coarse to fine-grained levels \cite{wen2021-CLVC-NET, jandial2021SAC}.

\textbf{Low-level and High-level.} 
SAC \cite{jandial2021SAC} 
focuses on the importance of both pixel and text and addresses the challenge of Text-Conditioned Image Retrieval (TCIR) through a two-step process. 
Trace  \cite{2020-TRACE} introduces a hierarchical feature aggregation module to learn composite vision-linguistic representations,  which can be considered a variant of SAC.
The multi-turn framework \cite{yuan2021multiturn} comprises three modules: the composite representation module,  the comparative analysis module,  and the fashion attribute module in order to learn composite representations. 
HCL \cite{HCL} encodes the images into three level representations (e.g. global,  entity and structure) and then fuses it through hierarchical composition learning.
In VAL \cite{Chen2020VAL},  a composite transformer that can be seamlessly integrated into a CNN is utilized to selectively preserve and transform visual features based on language semantics. 
Similar to VAL,  DATIR \cite{DATIR} also composes features in multiple levels,  which is a Distribution-Aligned Text-based Image Retrieval (DATIR) model,  incorporating attention mutual information maximization and hierarchical mutual information maximization. 

\textbf{Content and Style.} 
 \cite{Leveraging, Lee2021CoSMo} divide the features into content and  style features. Chawla et al. \cite{Leveraging} propose to represent the image using its style and content components,  then transform each of these components individually and merge for retrieval. 
CoSMo \cite{Lee2021CoSMo} stands for Content-Style Modulator,   founded on the concept of separately adjusting the content and style of a reference image in terms of the specified text. 
Uncertainty \cite{chen2024uncertainty} focuses on the alignment of coarse-grained retrieval by considering the multi-grained uncertainty. It integrates fine- and coarse-grained retrieval as matching data points with small and large fluctuations respectively and further proposes uncertainty modeling. 

\textbf{Global and Local.} 
In CLVCNet \cite{wen2021-CLVC-NET},  a Comprehensive Linguistic-Visual Composition Network is introduced,  which can effectively integrates both local-wise and global-wise compositions and achieve better results with a mutual enhancement mechanism.
In  \cite{peng2018cmgans},  the adversarial learning is introduced into TGIR in TIS \cite{TIS} task to learn discriminative representations of the query,  i.e. composite of reference image and the modification text by jointly training a generative model. 
FashionVLP \cite{Goenka_2022_FashionVLP} utilizes CNN as image feature extractor,  BERT for text encoder and VinVL \cite{zhang2021vinvl},  a Multilayer Vision-Language multimodal Transformer with Self-attention Mechanism,  for fusion.

    

\subsubsection{Transformer-based Methods}\label{Transformer based methods}

Transformers \cite{vaswani2023-Transformer,  devlin2019BERT} and their variants  \cite{zhao2022-PL4CIR_PLHMQ-twostage, CLIP4CIR2, baldrati2022combiner, Wen_2023-TG-CIR, baldrati2022-CLIP4CIR, CLIP4CIR3, liu2023BLIP4CIR1, liu2024-BLIP4CIR2, bai2023-SPRC-sentencelevel, ComqueryFormer, wu2020fashioniq, Goenka_2022_FashionVLP, tian2022AACL, levy2023CASE, pang2022MCR} have profoundly advanced the field of feature learning,  due to the global self-attention mechanism in modeling long-range dependence. 
Vaswani et al.  \cite{vaswani2023-Transformer} firstly features an encoder-decoder structure based on Transformer,  equipped with multi-head self-attention layers. 
This configuration is adept at learning contextual relationships within the data effectively.    
Compared with ResNet,  vision transformer (ViT) \cite{dosovitskiy2021-viT-g/14} and Swin Transformer \cite{liu2021swin-transformer} have stronger representational capability due to larger pre-training data,  which facilitate the generalization of the model for unknown distributions.
With the transformer architecture,  some methods \cite{tian2022AACL, ComqueryFormer} adopt the Swin Transformer \cite{liu2021swin-transformer} to encapsulate the visual feature and show a great potential to outperform CNN-based architecture in many vision tasks. Compared with other vision Transformers,  it can construct hierarchical image features and has linear computational complexity to image size.
AACL \cite{tian2022AACL} features an additive self-attention layer to selectively preserve and transform multi-level visual features conditioned on text semantics,  in order to derive an expressive composite representation.
ComqueryFormer \cite{ComqueryFormer} utilizes a cross-modal transformer as the traditional composition module. 
By dividing the query text into modification and auxiliary types,  CRN \cite{2023-CRN} is a Hierarchical Aggregation Transformer for Cross Relation learning,  towards 
relation-aware representation. 

    

    

\begin{table*}[h]
\centering
\caption{Other multimodal composite retrieval methods.}\label{table methods OMCR}
\fontsize{3}{3}\selectfont
\resizebox{\linewidth}{!}{
\begin{tabular}{llcccc}
\hline
\textbf{Methods}
& \textbf{Related modalities}
& \textbf{Encoder}
& \textbf{Composition}
& \textbf{Year}
\\
\hline
T2I-Adapter \cite{mou2023-T2I-Adapter}
& text,  sketch,  keypose
& CLIP-L
& Addition
& 2024 
\\
TWPW \cite{changpinyo2021-TWPW}
& text,  mouse trace
& CNN,  Transformer
& Concatenation
& 2021 
\\
MMFR \cite{2022-MMFR}
& text,  image,  audio
& BERT,  ResNet,  VGGish \cite{hershey2017-VGGISH}
& -
& 2022 
\\
SIMC \cite{2003-SIMC}
& text,  audio,  video
& GMM,  HMM,  SVM
& Bayesian Networks,  SVMs
& 2003
\\
LAVIMO \cite{yin2024-LAVIMO}
& text,  video,  motion
& DistilBERT \cite{sanh2020-DistilBERT},  CLIP,  MotionCLIP \cite{tevet2022-MotionCLIP}
& Attention Mechanism
& 2024
\\
TriCoLo \cite{ruan2023-TriCoLo}
& text,  image,  3D
& Bi-GRU,  MVCNN,  3D-CNN
& -
& 2024 
\\
    \hline
    \end{tabular}}
\end{table*}

\begin{table}[h]
\centering
\caption{The Publicly Available Datasets of Image-text Composite Editing.}
\renewcommand{\arraystretch}{0.9}
\resizebox{\linewidth}{!}{
\begin{tabular}{lccc}

\hline
\textbf{Dataset}
& \textbf{Modalities}
& \textbf{Scale}
& \textbf{Link}\\
\hline
\multicolumn{4}{c}{\selectfont \textbf{Image-text Composite Editing}}
\label{Datasets-editing}\\

\hline
CUB \cite{wah2011caltech-200-bird}
& Image,  Text      
& 11K images,  11K texts
&\href{https://www.vision.caltech.edu/datasets/cub_200_2011/}{\color{magenta}\Checkmark}
\\
Oxford-102 flower \cite{Oxford-102-flower}
& Image,  Text 
& 8K images,  8K texts
& \href{https://www.robots.ox.ac.uk/~vgg/data/flowers/102/}{\color{magenta}\Checkmark}
\\
CelebA \cite{liu2015-CelebA}
& Image,  Text 
& 202K images,  8M texts
& \href{https://mmlab.ie.cuhk.edu.hk/projects/CelebA.html}{\color{magenta}\Checkmark}
\\
Fashion Synthesis \cite{zhu2017prada}
& Image,  Text 
& 78K images,  -
& \href{https://mmlab.ie.cuhk.edu.hk/projects/DeepFashion/FashionSynthesis.html}{\color{magenta}\Checkmark}
\\
MIT-Adobe 5k \cite{MIT-Adobe}
& Image,  Text 
& 5K images,  20K texts
& \href{https://data.csail.mit.edu/graphics/fivek/}{\color{magenta}\Checkmark}
\\
MS-COCO \cite{lin2015-mscoco}
& Image,  Text 
& 164K images,  616K texts
& \href{https://cocodataset.org/}{\color{magenta}\Checkmark}
\\
ReferIt \cite{kazemzadeh-etal-2014-referit}
& Image,  Text 
& 19K images,  130K texts 
& \href{https://github.com/lichengunc/refer}{\color{magenta}\Checkmark}
\\
CLEVR \cite{johnson2016clevr}
& Image,  Text 
& 100K images,  865K texts
& \href{https://cs.stanford.edu/people/jcjohns/clevr/}{\color{magenta}\Checkmark}
\\
i-CLEVR \cite{elnouby2019-GeNeVA-GAN}
& Image,  Text 
& 10K sequences,  50K texts
& \href{https://github.com/topics/i-clevr}{\color{magenta}\Checkmark}
\\
CSS \cite{vo2018TIRG}
& Image,  Text 
& 34K images,  -
& \href{https://drive.google.com/file/d/1wPqMw-HKmXUG2qTgYBiTNUnjz83hA2tY/view?pli=1}{\color{magenta}\Checkmark}
\\
CoDraw \cite{kim2019codraw}
& Image,  Text 
& 9K images,  -
& \href{https://github.com/facebookresearch/CoDraw}{\color{magenta}\Checkmark}
\\
Cityscapes \cite{cordts2016cityscapes}
& Image,  Text 
& 25K images,  -
& \href{https://www.cityscapes-dataset.com/}{\color{magenta}\Checkmark}
\\
Zap-Seq \cite{cheng2020-SeqAttnGAN}
& Image,  Text 
& 8K images,  18K texts
& -
\\
DeepFashion-Seq \cite{cheng2020-SeqAttnGAN}
& Image,  Text 
& 4K images,  12K texts
& -
\\
FFHQ \cite{karras2019-StyleGAN}
& Image
& 70K images
& \href{https://github.com/NVlabs/ffhq-dataset}{\color{magenta}\Checkmark}
\\
LSUN \cite{yu2016-LSUN}
& Image
& 1M images
& \href{https://github.com/fyu/lsun/blob/master/README.md}{\color{magenta}\Checkmark}
\\
AFHQ  \cite{choi2020-AFHQ}
& Image
& 15K images
& \href{https://www.kaggle.com/datasets/andrewmvd/animal-faces/code}{\color{magenta}\Checkmark}
\\
CelebA-HQ \cite{karras2018-CelebA-HQ}
& Image
& 30K images
& \href{https://github.com/tkarras/progressive_growing_of_gans}{\color{magenta}\Checkmark}
\\
Animal faces \cite{animal-face}
& Image
& 16K images
& \href{https://www.kaggle.com/datasets/andrewmvd/animal-faces}{\color{magenta}\Checkmark}
\\
Landscapes \cite{landscapes}
& Image
& 4K images
& \href{https://www.kaggle.com/datasets/arnaud58/landscape-pictures}{\color{magenta}\Checkmark}
\\
\hline
\multicolumn{4}{c}{\textbf{Image-text Composite Retrieval}}

\label{Datasets-retrieval}
\\
\hline
Fashion200k \cite{han2017fashion200k}
& Image,  Text
& 200K images,  200K texts
& \href{https://github.com/xthan/fashion-200k}{\color{magenta}\Checkmark}
\\
MIT-States \cite{isola2015MIT-States}
& Image,  Text
& 53K images,  53K texts
& \href{https://web.mit.edu/phillipi/Public/states_and_transformations/index.html}{\color{magenta}\Checkmark}
\\
Fashion IQ \cite{wu2020fashioniq}
& Image,  Text
& 77K images,  -
& \href{https://github.com/XiaoxiaoGuo/fashion-iq}{\color{magenta}\Checkmark}
\\
CIRR \cite{liu2021CIRPLANT}
& Image,  Text
& 21K images,  - 
& \href{https://github.com/Cuberick-Orion/CIRR}{\color{magenta}\Checkmark}
\\
CSS \cite{vo2018TIRG}
& Image,  Text
& 34K images,  -
& \href{https://drive.google.com/file/d/1wPqMw-HKmXUG2qTgYBiTNUnjz83hA2tY/view?pli=1}{\color{magenta}\Checkmark}
\\
Shoes \cite{Berg2010-Shoes}
& Image
& 14K images
& \href{https://www.kaggle.com/datasets/noobyogi0100/shoe-dataset}{\color{magenta}\Checkmark}
\\
Birds-to-Words \cite{forbes2019-Birds-to-Words}
& Image,  Text
& -
& \href{https://github.com/google-research-datasets/birds-to-words}{\color{magenta}\Checkmark}
\\
SketchyCOCO \cite{gao2020-sketchycoco}
& Image,  Sketche
& 14K sketches,  14K images
& \href{https://github.com/sysu-imsl/SketchyCOCO}{\color{magenta}\Checkmark}
\\
FSCOCO \cite{chowdhury2022fscoco}
& Image,  Sketche
& 10K sketches
& \href{https://www.pinakinathc.me/fscoco/}{\color{magenta}\Checkmark}
\\
 \hline
\multicolumn{4}{c}{\textbf{Other Multimodal Composite Retrieval}}\label{Datasets-other}
\\
\hline


\hline
HumanML3D \cite{guo2022-HumanML3D}
& Motion,  Text
& 14K motion sequences,  44K texts
&   \href{https://github.com/EricGuo5513/HumanML3D}{\color{magenta}\Checkmark}
\\
KIT-ML \cite{plappert2016-kit-ML}
& Motion,  Text
& 3K motions,  6K texts
&    \href{https://h2t.iar.kit.edu/english/1445.php}{\color{magenta}\Checkmark}
\\
Text2Shape \cite{chen2019-text2shape}
& Shape,  Text
& 6K chairs,  8K tables,  70K texts
&  \href{https://github.com/kchen92/text2shape}{\color{magenta}\Checkmark}  
\\
Flickr30k LocNar \cite{plummer2016flickr30k}
& Image,  Text
& 31K images,  155K texts
&  \href{https://www.kaggle.com/datasets/hsankesara/flickr-image-dataset}{\color{magenta}\Checkmark}  
\\
Conceptual Captions \cite{sharma2018-conceptual-captions}
& Image,  Text
& 3.3M images,  33M texts
&  \href{https://github.com/google-research-datasets/conceptual-captions}{\color{magenta}\Checkmark}  
\\
Sydney\_IV \cite{mao2018-RS-dataset}
& Image,  Audio
& 613 images,  3K audio
& \href{https://github.com/201528014227051/RSICD_optimal}{\color{magenta}\Checkmark}   
\\
UCM\_IV \cite{mao2018-RS-dataset}
& Image,  Audio
& 2K images,  10K audio
& \href{https://github.com/201528014227051/RSICD_optimal}{\color{magenta}\Checkmark}   
\\
RSICD\_IV \cite{mao2018-RS-dataset}
& Image,  Audio
& 11K images,  55K audio
& \href{https://github.com/201528014227051/RSICD_optimal}{\color{magenta}\Checkmark}   
\\
\hline

\end{tabular}}
\label{datasets}
\end{table}

\subsubsection{VLP-based Methods}\label{Large Model based methods}

    As illustrated in Table \ref{ITCR methods},  Vision-Language Pre-training (VLP) methods harness the power of large-scale Vision-Language Models (VLMs) such as CLIP and BLIP to improve the retrieval performance by introducing large amount of pre-trained data. For clarity,  
    we classify the VLP based ITCR methods into Fusion based and Inverse based. 
    In fusion based methods,  the encoder of VLMs are utilized to extract better feature representation. In inverse based methods,  the VLMs are used to generate detailed descriptions of reference images. For instance,  BLIP excels at producing comprehensive textual representations by using large language models (LLMs) like Llama and GPT. These LLMs integrate the generated descriptions with specific textual queries to craft precise captions for the target images. Subsequently,  the VLM CLIP plays a critical role in retrieving the target image by leveraging its sophisticated ability to comprehend and associate visual and textual information.
    This innovation offers enhanced flexibility in addressing composite image retrieval challenges. Certain methods \cite{zhao2022-PL4CIR_PLHMQ-twostage, CLIP4CIR2, baldrati2022combiner, Wen_2023-TG-CIR, baldrati2022-CLIP4CIR} have demonstrated success in solving the image retrieval task by fine-tuning CLIP \cite{radford2021CLIP} for improved performance.
    
\textbf{Fusion-based methods}. Contrastive Language-Image Pre-training (CLIP)  \cite{radford2021CLIP} is pre-trained on a large-scale dataset of image and text pairs from open domains,  demonstrating exceptional capability in image retrieval tasks. By using 400 million image-text pairs scraped from the web,  CLIP learns associations between images and their textual descriptions. It uses two independent encoders to achieve superior retrieval results in open-domain image-text retrieval tasks. 
 \cite{zhao2022-PL4CIR_PLHMQ-twostage,  baldrati2022-CLIP4CIR,  baldrati2022combiner} leverage the open-domain semantic joint embedding space based on the pre-trained CLIP foundation model.    
CLIP4CIR \cite{baldrati2022-CLIP4CIR} slightly modifies the architecture of the Combiner network depicted in  \cite{CLIP4CIR2}.  TG-CIR \cite{Wen_2023-TG-CIR} is a Target-Guided composite Image Retrieval network,  which includes the CLIP module for image-text embedding and a multimodal query composition module guided by the target-query relationship. 
PL4CIR \cite{zhao2022-PL4CIR_PLHMQ-twostage} introduces a multi-stage progressive learning framework based on CLIP and a self-supervised query adaptive composite module. 
CompoDiff \cite{gu2024compodiff} combines CLIP and Diffusion with a mask by employing a diffusion process in the frozen CLIP latent feature space with classifier-free guidance (CFG) and adopting the Transformer architecture for the denoiser.
    
In  \cite{chen2023ranking-aware},  a ranking-aware uncertainty approach is proposed for image-text composite retrieval,  which incorporates in-sample uncertainty 
,  cross-sample uncertainty 
and distribution regularization to align the feature distributions of the target and source.
PALAVRA \cite{cohen2022-PALAVRA} employs a two-stage approach based on textual inversion. It begins with a pre-trained mapping function and is followed by an optimization process aimed at the pseudo-word token,  which encodes object sets into CLIP's textual embedding space.
Based on the architecture in  \cite{baldrati2022-CLIP4CIR},  in BLIP4CIR \cite{liu2023BLIP4CIR1, liu2024-BLIP4CIR2, baldrati2022-CLIP4CIR},  a candidate re-ranking model featuring a dual-encoder architecture and a bidirectional training approach are proposed. 
In SPRC \cite{bai2023-SPRC-sentencelevel},  a pre-trained VLM,  e.g.,  BLIP-2,  is leveraged to generate sentence-level prompts for the relative caption,  towards 
text-to-image retrieval module.    
CASE \cite{levy2023CASE} introduces a novel baseline that leverages pre-trained BLIP components with early fusion,  named Cross-Attention driven Shift Encoder (CASE). It comprises two transformer components: BERT based shift encoder and a ViT encoder. 

\textbf{Inverse-based methods}.
The standard ITCR task usually needs the triplets $(I_r, T_r, I_t)$ comprised of a reference image,  a modified text,  and a target image respectively. Unlike pair-wise image-text dataset (e.g. CC3M \cite{sharma2018-CC3M} and LAION \cite{schuhmann2021laion400m}),  it cannot be easily crawled from the Internet and needs human-label to describe the $T_r$ between the image-text pair \cite{liu2021CIRPLANT, wu2020fashioniq},  which is costly and time-consuming.

Although previous methods for composite image retrieval have demonstrated encouraging outcomes,  their dependence on costly manually-annotated datasets constrains their scalability and limits their applicability across various domains distinct from those of the training datasets. To overcome this,  zero-shot method based text inversion attracts much attention since 2023. Zero-shot learning in  composite image retrieval  represents a cutting-edge approach where the model is designed to generalize to new tasks without needing explicit examples during training. These models leverage vast amounts of unlabelled data and inherent knowledge captured during pre-training on diverse tasks. In a zero-shot scenario,  the alignment is typically facilitated by embedding both text and image features into a shared semantic space where alignment does not rely on direct feature fusion but the semantic consistency between the modalities.
Specifically,  numerous studies \cite{saito2023pic2word, Baldrati2023SEARLE, tang2023contexti2w, chen2023-PLI, gu2024LinCIR} indicate a growing interest in refining image retrieval techniques to enhance their efficiency and domain adaptability without the reliance on extensive annotated resources. Zero-shot models are particularly advantageous in scenarios where annotated data is scarce or when the task requires the model to understand and align novel concepts that were not present in the training data. They do not require training data and built on pre-trained VLMs (e.g. CLIP). However,  there is still gap between the pre-trained task and image-text composite retrieval.  PLI \cite{chen2023-PLI} uses a novel mask tuning self-supervised pre-training approach in order to reduce the gap. By randomly masking the original reference image and using text input to reconstruct the original unmasked image semantics,  the pre-tuning process can minimize the similarity between the query and the target. 

\begin{table*}
\centering
\large
\caption{Performance comparison on the Fashion-IQ dataset  \cite{wu2020fashioniq} (\underline{VAL split}). Notably,  R@K refers to Recall rate for top K. 
A higher value indicates better performance.}
\fontsize{5}{5}\selectfont
\renewcommand{\arraystretch}{0.9}
\resizebox{\linewidth}{!}{
\begin{tabular}{ll|cc|cc|cc|cc|c}
\hline
\label{Fashion-IQ baseline}
         \multirow{2}{*}{\textbf{Methods}}& \multirow{2}{*}{\textbf{Image encoder}}& \multicolumn{2}{c|}{\textbf{Dress}}& \multicolumn{2}{c|}{\textbf{Shirt}}&  \multicolumn{2}{c|}{ \textbf{Toptee}}&  \multicolumn{2}{c|}{\textbf{Average}}&\multirow{2}{*}{\textbf{Avg}. }\\
         & & R@10&  R@50&  R@10&  R@50&  R@10&  R@50&  R@10&  R@50\\
\hline
        ARTEMIS+LSTM \cite{delmas2022ARTEMIS}&  ResNet-18&  25.23&  48.64&  20.35&  43.67&  23.36&  446.97&  22.98&  46.43& 34.70\\
        ARTEMIS+BiGRU \cite{delmas2022ARTEMIS}&  ResNet-18&  24.84&  49.00&  20.40&  43.22&  23.63&  47.39&  22.95&  46.54& 34.75\\
        JPM(VAL, MSE) \cite{JPM}&  ResNet-18&  21.27&  43.12&  21.88&  43.3&  25.81&  50.27& 22.98&  45.59& 34.29\\
        JPM(VAL, Tri) \cite{JPM}&  ResNet-18&  21.38&  45.15&  22.81&  45.18&  27.78&  51.70& 23.99&  47.34& 35.67\\
        EER \cite{EER}&  ResNet-50&  30.02&  55.44&  25.32&  49.87&  33.20&  60.34& 29.51& 55.22& 42.36\\
        Ranking-aware \cite{chen2023ranking-aware}& ResNet-50 &  34.80&  60.22&  45.01&  69.06&  47.68&  74.85& 42.50& 68.04& 55.27\\     
        CRN \cite{2023-CRN}& ResNet-50 &  30.20&  57.15&  29.17&  55.03&  33.70&  63.91& 31.02& 58.70& 44.86\\
        DWC \cite{huang2023-DWC}& ResNet-50& 32.67& 57.96& 35.53& 60.11& 40.13& 66.09& 36.11& 61.39& 48.75\\
        DATIR \cite{zhao2022-PL4CIR_PLHMQ-twostage}& ResNet-50 &  21.90&  43.80&  21.90&  43.70&  27.20&  51.60& 23.70&  46.40& 35.05\\
        CoSMo \cite{AMC}&  ResNet-50&  25.64&  50.30&  24.90&  49.18&  29.21&  57.46&  26.58&  52.31& 39.45\\
        FashionVLP \cite{ComqueryFormer}&  ResNet-50&  32.42&  60.29&  31.89&  58.44&  38.51&  68.79& 34.27&  62.51& 48.39\\
        CLVC-Net \cite{wen2021-CLVC-NET}&  ResNet-50& 29.85& 56.47& 28.75& 54.76& 33.50& 64.00&  30.70& 58.41& 44.56 \\
        SAC w/BERT \cite{jandial2021SAC}&  ResNet-50&  26.52&  51.01&  28.02&  51.86&  32.70&  61.23&  29.08&  54.70& 41.89\\
        SAC w/ Random Emb. \cite{jandial2021SAC}& ResNet-50& 26.13& 52.10& 26.20& 50.93& 31.16& 59.05& 27.83& 54.03& 40.93\\
        DCNet \cite{kim2021-DCNet}&  ResNet-50&  28.95&  56.07&  23.95&  47.30&  30.44&  58.29&  27.78&  53.89& 40.83\\
        AMC \cite{AMC}&  ResNet-50&  31.73&  59.25&  30.67&  59.08&  36.21&  66.60&  32.87&  61.64& 47.25\\
        VAL($\mathcal{L}_{vv}$) \cite{Chen2020VAL}&  ResNet-50& 21.12& 42.19& 21.03& 43.44& 25.64& 49.49&  22.60& 45.04&  33.82\\
        ARTEMIS+LSTM \cite{delmas2022ARTEMIS}&  ResNet-50& 27.34 &  51.71&  21.05&  44.18&  24.91&  49.87&  24.43&  48.59& 36.51\\
        ARTEMIS+BiGRU \cite{delmas2022ARTEMIS}&  ResNet-50&  27.16&  52.40&  21.78&  43.64&  29.20&  54.83&  26.05&  50.29& 38.17\\
        VAL($\mathcal{L}_{vv}$ + $\mathcal{L}_{vs}$) \cite{Chen2020VAL}&  ResNet-50& 21.47& 43.83& 21.03& 42.75& 26.71& 51.81&  23.07& 46.13& 34.60 \\
        VAL(GloVe)&  ResNet-50&  22.53&  44.00&  22.38&  44.15&  27.53&  51.68&  24.15&  46.61& 35.38\\
        AlRet \cite{xu2024-AlRet}& ResNet-50& 30.19& 58.80& 29.39& 55.69& 37.66& 64.97& 32.36& 59.76& 46.12\\
        RTIC \cite{shin2021RTIC}&  ResNet-50& 19.40& 43.51& 16.93& 38.36& 21.58& 47.88& 19.30& 43.25& 31.28\\
        RTIC-GCN \cite{shin2021RTIC}&  ResNet-50& 19.79& 43.55& 16.95&  38.67& 21.97& 49.11& 19.57& 43.78&  31.68\\
        Uncertainty (CLVC-Net) \cite{chen2024uncertainty}& ResNet-50 & 30.60&  57.46&  31.54&  58.29&  37.37&  68.41&  33.17&  61.39& 47.28\\
        Uncertainty (CLIP4CIR) \cite{chen2024uncertainty}& ResNet-50 & 32.61&  61.34&  33.23&  62.55&  41.40&  72.51&  35.75&  65.47& 50.61\\ 
        CRR \cite{CRR}&  ResNet-101&  30.41&  57.11&  33.67&  64.48&  30.73&  58.02& 31.60&  59.87& 45.74\\ 
        CIRPLANT \cite{liu2021CIRPLANT}&  ResNet-152&  14.38&  34.66&  13.64&  33.56&  16.44&  38.34&  14.82&  35.52& 25.17\\
        CIRPLANT w/OSCAR \cite{liu2021CIRPLANT}&  ResNet-152&  17.45&  40.41&  17.53&  38.81&  21.64&  45.38&  18.87&  41.53& 30.20\\
        ComqueryFormer \cite{ComqueryFormer}& Swin & 33.86&  61.08&  35.57&  62.19&  42.07&  69.30& 37.17&  64.19& 50.68\\
        CRN \cite{2023-CRN}& Swin  &  30.34&  57.61&  29.83&  55.54&  33.91&  64.04& 31.36& 59.06& 45.21\\
        CRN \cite{2023-CRN}& Swin-L &  32.67&  59.30&  30.27&  56.97&  37.74&  65.94& 33.56& 60.74& 47.15\\
        BLIP4CIR1 \cite{liu2023BLIP4CIR1}&  BLIP-B&  43.78&  67.38&  45.04&  67.47&  49.62&  72.62&  46.15&  69.15& 57.65\\
        CASE \cite{levy2023CASE}&  BLIP&  47.44&  69.36&  48.48& 70.23&  50.18&  72.24&  48.79&  70.68& 59.74\\
        BLIP4CIR2 \cite{liu2024-BLIP4CIR2}&  BLIP&  40.65&  66.34&  40.38&  64.13&  46.86&  69.91&  42.63&  66.79& 54.71\\
        BLIP4CIR2+Bi \cite{liu2024-BLIP4CIR2}&  BLIP&  42.09&  67.33&  41.76&  64.28&  46.61&  70.32&  43.49&  67.31& 55.40\\
        CLIP4CIR3 \cite{CLIP4CIR3}& CLIP&  39.46&  64.55&  44.41&  65.26&  47.48&  70.98& 43.78& 66.93& 55.36\\ 
        CLIP4CIR \cite{CLIP4CIR2}&  CLIP&  33.81&  59.40&  39.99&  60.45&  41.41&  65.37&  38.32&  61.74& 50.03\\
        AlRet \cite{xu2024-AlRet}& CLIP-RN50& 40.23& 65.89& 47.15& 70.88& 51.05& 75.78& 46.10& 70.80& 58.50\\ 
        Combiner \cite{baldrati2022combiner}& CLIP-RN50& 31.63& 56.67& 36.36& 58.00& 38.19& 62.42& 35.39& 59.03& 47.21\\
        DQU-CIR \cite{Wen_2024-DQU-CIR}& CLIP-H &57.63& 78.56& 62.14& 80.38& 66.15& 85.73& 61.97& 81.56& 71.77\\
        PL4CIR \cite{zhao2022-PL4CIR_PLHMQ-twostage}& CLIP-L &  38.18&  64.50&  48.63&  71.54&  52.32&  76.90& 46.37&  70.98& 58.68\\
        TG-CIR \cite{Wen_2023-TG-CIR}&  CLIP-B &  45.22&  69.66&  52.60&  72.52&  56.14&  77.10& 51.32& 73.09& 62.21\\
        PL4CIR \cite{zhao2022-PL4CIR_PLHMQ-twostage}& CLIP-B &  33.22&  59.99&  46.17&  68.79&  46.46&  73.84& 41.98&  67.54& 54.76\\
\hline
\end{tabular}}
\end{table*}

\begin{table*}
\centering
\large
\caption{Performance comparison on the Fashion-IQ dataset  \cite{wu2020fashioniq} (\underline{original split}). Notably,  R@K refers to Recall rate for top K. 
A higher value indicates better performance.}
\fontsize{5}{5}\selectfont
\renewcommand{\arraystretch}{0.9}
\resizebox{\linewidth}{!}{
\begin{tabular}{ll|cc|cc|cc|cc|c}
\hline
\label{Fashion-IQ baseline 2}
\multirow{2}{*}{\textbf{Methods}}& \multirow{2}{*}{\textbf{Image encoder}}& \multicolumn{2}{c|}{\textbf{Dress}}& \multicolumn{2}{c|}{\textbf{Shirt}}&  \multicolumn{2}{c|}{ \textbf{Toptee}}&  \multicolumn{2}{c|}{\textbf{Average}}&\multirow{2}{*}{\textbf{Avg}. }\\

         &  & R@10&  R@50&  R@10&  R@50&  R@10&  R@50&  R@10&  R@50\\         
\hline
         ComposeAE \cite{anwaar2021composeAE}&  ResNet-18& 10.77& 28.29& 9.96& 25.14& 12.74& 30.79& - &  - & -\\
         TIRG \cite{vo2018TIRG}&  ResNet-18&  14.87&  34.66&  18.26&  37.89&  19.08&  39.62&  17.40&  37.39& 27.40\\
         MAAF \cite{AMC}&  ResNet-50&  23.80&  48.60&  21.30&  44.20&  27.90&  53.60&  24.30&  48.80& 36.60\\
         Leveraging \cite{Leveraging}&  ResNet-50 &  19.33&  43.52&  14.47&  35.47&  19.73&  44.56& 17.84&  41.18& 29.51\\
         MCR \cite{pang2022MCR}&  ResNet-50&  26.20&  51.20&  22.40&  46.01&  29.70&  56.40& 26.10&  51.20& 38.65\\
         MCEM ($\mathcal{L}_{\tiny CE}$) \cite{MCEM}& ResNet-50& 30.07& 56.13& 23.90& 47.60& 30.90& 57.52& 28.29& 53.75& 41.02\\
         MCEM ($\mathcal{L}_{\small FCE}$) \cite{MCEM}& ResNet-50& 31.50& 58.41& 25.01& 49.73& 32.77& 61.02& 29.76& 56.39& 43.07\\
         MCEM ($\mathcal{L}_{\small AFCE}$) \cite{MCEM}& ResNet-50& 33.23& 59.16& 26.15& 50.87& 33.83& 61.40& 31.07& 57.14& 44.11\\
         AlRet \cite{xu2024-AlRet}& ResNet-50& 27.34& 53.42& 21.30& 43.08& 29.07& 54.21& 25.86& 50.17& 38.02\\ 
         MCEM ($\mathcal{L}_{\small AFCE}$ w/ BERT) \cite{MCEM}& ResNet-50& 32.11& 59.21& 27.28& 52.01& 33.96& 62.30& 31.12& 57.84& 44.48\\       
         JVSM \cite{chen2020JVSM}&  MobileNet-v1& 10.70& 25.90& 12.00& 27.10& 13.00& 26.90&11.90&26.63 & 19.27 \\
         FashionIQ(Dialog Turn 1) \cite{wu2020fashioniq}& EfficientNet-b& 12.45& 35.21& 11.05& 28.99& 11.24& 30.45& 11.58& 31.55& 21.57\\
         FashionIQ(Dialog Turn 5) \cite{wu2020fashioniq}& EfficientNet-b& 41.35& 73.63& 33.91& 63.42& 33.52& 63.85& 36.26& 66.97& 51.61\\
         AACL \cite{tian2022AACL}& Swin &  29.89&  55.85&  24.82&  48.85&  30.88&  56.85&  28.53&  53.85& 41.19\\
         ComqueryFormer  \cite{ComqueryFormer}& Swin & 28.85& 55.38& 25.64& 50.22& 33.61& 60.48& 29.37& 55.36& 42.36\\
         AlRet \cite{xu2024-AlRet}& CLIP & 35.75& 60.56& 37.02& 60.55& 42.25& 67.52& 38.30& 62.82& 50.56\\
         MCEM ($\mathcal{L}_{\small AFCE}$)  \cite{MCEM}& CLIP & 33.98& 59.96& 40.15& 62.76& 43.75& 67.70& 39.29& 63.47& 51.38\\ 
         SPN (TG-CIR) \cite{feng2024data_generation-SPN}&CLIP&  36.84&  60.83&  41.85&  63.89&  45.59&  68.79&  41.43&  64.50&  52.97\\
         SPN (CLIP4CIR) \cite{feng2024data_generation-SPN}&CLIP&  38.82&  62.92&  45.83&  66.44&  48.80&  71.29&  44.48&  66.88&  55.68\\
         PL4CIR \cite{zhao2022-PL4CIR_PLHMQ-twostage}& CLIP-B &  29.00&  53.94&  35.43&  58.88&  39.16&  64.56& 34.53& 59.13& 46.83\\
         FAME-ViL \cite{han2023FAMEvil}& CLIP-B &  42.19&  67.38&  47.64&  68.79&  50.69&  73.07& 46.84& 69.75& 58.30\\
         PALAVRA \cite{cohen2022-PALAVRA}&  CLIP-B&  17.25&  35.94&  21.49&  37.05&  20.55&  38.76& 19.76& 37.25& 28.51\\
         MagicLens-B \cite{zhang2024magiclens}&  CLIP-B &  21.50&  41.30&  27.30&  48.80&  30.20&  52.30& 26.30& 47.40& 36.85\\
         SEARLE \cite{Baldrati2023SEARLE}&  CLIP-B &  18.54&  39.51&  24.44&  41.61&  25.70&  46.46& 22.89& 42.53& 32.71\\
         CIReVL \cite{karthik2024-CIReVL}& CLIP-B & 25.29& 46.36& 28.36& 47.84& 31.21& 53.85& 28.29& 49.35& 38.82\\
         SEARLE-OTI \cite{Baldrati2023SEARLE}&  CLIP-B &  17.85&  39.91&  25.37&  41.32&  24.12&  45.79& 22.44&  42.34& 32.39\\
         PLI \cite{chen2023-PLI}&  CLIP-B &  25.71&  47.81&  33.36&  53.47&  34.87&  58.44& 31.31& 53.24& 42.28\\
         PL4CIR \cite{zhao2022-PL4CIR_PLHMQ-twostage}& CLIP-L &  33.60&  58.90&  39.45&  61.78&  43.96&  68.33& 39.02& 63.00& 51.01\\
         SEARLE-XL \cite{Baldrati2023SEARLE}& CLIP-L &  20.48&  43.13&  26.89&  45.58&  29.32&  49.97& 25.56& 46.23& 35.90\\
         SEARLE-XL-OTI \cite{Baldrati2023SEARLE}& CLIP-L  &  21.57&  44.47&  30.37&  47.49&  30.90&  51.76& 27.61& 47.90& 37.76\\
         Context-I2W \cite{tang2023contexti2w}& CLIP-L &  23.10&  45.30&  29.70&  48.60&  30.60&  52.90& 27.80& 48.90& 38.35\\
         CompoDiff(with SynthTriplets18M) \cite{gu2024compodiff}&  CLIP-L &  32.24&  46.27&  37.69&  49.08&  38.12&  50.57& 36.02&  48.64& 42.33\\
         CompoDiff(with SynthTriplets18M) \cite{gu2024compodiff}&  CLIP-L &  37.78&  49.10&  41.31&  55.17&  44.26&  56.41& 39.02&  51.71& 46.85\\
         Pic2Word \cite{saito2023pic2word}&  CLIP-L&  20.00&  40.20&  26.20&  43.60&  27.90&  47.40& 24.70& 43.70& 34.20\\
         PLI \cite{chen2023-PLI}&  CLIP-L &  28.11&  51.12&  38.63&  58.51&  39.42&  62.68& 35.39& 57.44& 46.42\\
         KEDs \cite{suo2024KEDs}& CLIP-L & 21.70& 43.80& 28.90& 48.00& 29.90& 51.90& 26.80& 47.90& 37.35\\
         CIReVL \cite{karthik2024-CIReVL}& CLIP-L & 24.79& 44.76& 29.49& 47.40& 31.36& 53.65& 28.55& 48.57& 38.56\\
         LinCIR \cite{gu2024LinCIR}&  CLIP-L &  20.92&  42.44&  29.10&  46.81&  28.81&  50.18& 26.28& 46.49& 36.39\\
         MagicLens-L \cite{zhang2024magiclens}&  CLIP-L &  25.50&  46.10&  32.70&  53.80&  34.00&  57.70& 30.70& 52.50& 41.60\\
         LinCIR \cite{gu2024LinCIR}&  CLIP-H &  29.80&  52.11&  36.90&  57.75&  42.07&  62.52& 36.26& 57.46& 46.86\\
         DQU-CIR \cite{Wen_2024-DQU-CIR}& CLIP-H &51.90& 74.37& 53.57& 73.21& 58.48& 79.23& 54.65& 75.60& 65.13\\
        LinCIR \cite{gu2024LinCIR}&  CLIP-G &  38.08&  60.88&  46.76&  65.11&  50.48&  71.09& 45.11&65.69 & 55.40\\
        CIReVL \cite{karthik2024-CIReVL}& CLIP-G & 27.07& 49.53& 33.71& 51.42& 35.80& 56.14& 32.19& 52.36& 42.28\\
        
         MagicLens-B \cite{zhang2024magiclens}&  CoCa-B &  29.00&  48.90&  36.50&  55.50&  40.20&  61.90& 35.20& 55.40& 45.30\\
         MagicLens-L \cite{zhang2024magiclens}&  CoCa-L &  32.30&  52.70&  40.50&  59.20&  41.40&  63.00& 38.00& 58.20& 48.10\\
         SPN (BLIP4CIR1) \cite{feng2024data_generation-SPN}& BLIP&  44.52&  67.13&  45.68&  67.96&  50.74&  73.79&  46.98&  69.63& 58.30 \\
         PLI \cite{chen2023-PLI}&  BLIP-B & 28.62& 50.78& 38.09& 57.79& 40.92&  62.68& 35.88& 57.08& 46.48\\
         SPN (SPRC) \cite{feng2024data_generation-SPN}& BLIP-2&  50.57&  74.12&  57.70&  75.27&  60.84&  79.96&  56.37&  76.45&  66.41\\
         CurlingNet \cite{yu2020Curlingnet}& - &  24.44&  47.69&  18.59&  40.57&  25.19&  49.66&  22.74&  45.97& 34.36\\
\bottomrule
\end{tabular}}
\end{table*}

\begin{table}
    \centering
    \caption{Performance comparison on the Fashion200k  \cite{han2017fashion200k}. Notably,  R@K refers to Recall rate for top K. 
A higher value indicates better performance.}
    \renewcommand{\arraystretch}{0.9}
    \resizebox{\linewidth}{!}{
    \begin{tabular}{llccc}
    \hline
    \label{Fashion200k baseline}
         Method&  Image encoder& R@1  & R@10 & R@50 \\
    \hline
         TIRG \cite{vo2018TIRG}&  ResNet-18& 14.10&  42.50& 63.80\\
         ComposeAE \cite{anwaar2021composeAE}&  ResNet-18& 22.80& 55.30&  73.40\\
         HCL \cite{HCL}& ResNet-18& 23.48&  54.03&  73.71\\
         CoSMo \cite{Lee2021CoSMo}& ResNet-18 & 23.30& 50.40&  69.30\\
         JPM(TIRG, MSE) \cite{JPM}& ResNet-18&19.80& 46.50&  66.60\\
         JPM(TIRG, Tri) \cite{JPM}& ResNet-18& 17.70&44.70&  64.50\\
         ARTEMIS \cite{delmas2022ARTEMIS}& ResNet-18& 21.50 &  51.10&  70.50\\
         GA(TIRG-BERT) \cite{huang2022-GA-data-augmentation}& ResNet-18&31.40&54.10& 77.60\\
         LGLI \cite{huang2023-LGLI}& ResNet-18& 26.50& 58.60& 75.60\\
         AlRet \cite{xu2024-AlRet}& ResNet-18& 24.42&  53.93& 73.25\\
         FashionVLP \cite{ComqueryFormer}& ResNet-18&  -& 49.9&  70.5\\
         CLVC-Net \cite{wen2021-CLVC-NET}& ResNet-50& 22.6& 53.00 & 72.20\\
         Uncertainty \cite{chen2024uncertainty}& ResNet-50& 21.80& 52.10& 70.20\\
         MCR \cite{ComqueryFormer}& ResNet-50&  49.40& 69.40& 59.40\\
         CRN \cite{2023-CRN}& ResNet-50& -& 53.10& 73.00\\
         EER w/ Random Emb. \cite{EER}& ResNet-50&  -&51.09& 70.23\\
         EER w/ GloVe \cite{EER}& ResNet-50& -& 50.88& 73.40 \\
         DWC \cite{huang2023-DWC}& ResNet-50& 36.49&  63.58& 79.02\\
         JGAN \cite{JGAN}&  ResNet-101&17.34& 45.28&  65.65 \\
         CRR \cite{CRR}&ResNet-101&  24.85& 56.41&  73.56\\
         GSCMR \cite{2022-GSCMR}& ResNet-101&21.57&  52.84& 70.12\\         
         VAL(GloVe) \cite{Chen2020VAL}& MobileNet& 22.90&  50.80&  73.30\\
         VAL($\mathcal{L}_{vv}$+$\mathcal{L}_{vs}$) \cite{Chen2020VAL}& MobileNet& 21.50&  53.80&  72.70\\ 
         DATIR \cite{ComqueryFormer}& MobileNet & 21.50& 48.80& 71.60\\
         VAL($\mathcal{L}_{vv}$) \cite{Chen2020VAL}& MobileNet& 21.20& 49.00&  68.80\\
         JVSM \cite{chen2020JVSM}&  MobileNet-v1&19.00& 52.10&  70.00\\
         TIS \cite{TIS} & MobileNet-v1& 17.76&47.54& 68.02\\
         DCNet \cite{kim2021-DCNet}& MobileNet-v1&- &  46.89&  67.56\\ 
         TIS \cite{TIS}& Inception-v3& 16.25& 44.14&  65.02\\
         LBF(big) \cite{hosseinzadeh2020-locally-LBF}& Faster-RCNN&17.78& 48.35&  68.50\\
         LBF(small) \cite{hosseinzadeh2020-locally-LBF}& Faster-RCNN&16.26& 46.90& 71.73\\
         ProVLA \cite{Hu_2023_ICCV-ProVLA}& Swin & 21.70& 53.70& 74.60\\
         CRN \cite{2023-CRN}& Swin & -&  53.30& 73.30\\
         ComqueryFormer \cite{ComqueryFormer}&Swin &  - & 52.20&  72.20\\ AACL \cite{tian2022AACL}&Swin &  19.64&  58.85&  78.86\\
         CRN \cite{2023-CRN}& Swin-L &  - & 53.50&  74.50\\
         DQU-CIR \cite{Wen_2024-DQU-CIR}& CLIP-H& 36.80& 67.90& 87.80\\
    \hline
    \end{tabular}}
\end{table}

\begin{table}
    \centering
    \caption{Performance comparison on the MIT-States \cite{isola2015MIT-States} dataset. Notably,  R@K refers to Recall rate for top K. 
A higher value indicates better performance.}

\renewcommand{\arraystretch}{0.9}
    \resizebox{\linewidth}{!}{
    \begin{tabular}{llcccc}
    \hline
    \label{MIT-States baseline}
         Method& Image encoder&R@1 &  R@5 & R@10 & Average\\
    \hline
         TIRG \cite{vo2018TIRG}& ResNet-18 & 12.20&  31.90&  43.10& 29.10\\
         ComposeAE \cite{anwaar2021composeAE}&  ResNet-18& 13.90&  35.30&  47.90& 32.37\\
         HCL \cite{HCL}& ResNet-18& 15.22&  35.95&  46.71& 32.63\\
         GA(TIRG) \cite{huang2022-GA-data-augmentation}& ResNet-18&13.60& 32.40& 43.20&29.70\\
         GA(TIRG-BERT) \cite{huang2022-GA-data-augmentation}& ResNet-18&15.40& 36.30& 47.70& 33.20\\
         GA(ComposeAE) \cite{huang2022-GA-data-augmentation}& ResNet-18&14.60& 37.00& 47.90&33.20\\
         LGLI \cite{huang2023-LGLI}& ResNet-18& 14.90& 36.40& 47.70& 33.00\\
         MAAF \cite{dodds2020MAAF}& ResNet-50& 12.70& 32.60& 44.80& - \\
         MCR \cite{CRR}& ResNet-50& 14.30&  35.36&  47.12& 32.26\\
         CRR \cite{CRR}& ResNet-101& 17.71&  37.16&  47.83& 34.23\\
         JGAN \cite{JGAN}& ResNet-101& 14.27&  33.21&  45.34& 29.1\\
         GSCMR \cite{2022-GSCMR}& ResNet-101&17.28& -&36.45& -\\
         TIS \cite{TIS}& Inception-v3& 13.13&  31.94&  43.32& 29.46\\
         LBF(big) \cite{hosseinzadeh2020-locally-LBF}& Faster-RCNN&14.72&  35.30&  46.56& 96.58 \\
         LBF(small) \cite{hosseinzadeh2020-locally-LBF}& Faster-RCNN&14.29& -& 34.67& 46.06\\
    \hline
    \end{tabular}}
\end{table}

\begin{table}
    \centering
    \caption{Performance comparison on the CSS \cite{vo2018TIRG} dataset.  Notably,  R@K refers to Recall rate for top K. 
A higher value indicates better performance.}
    \resizebox{\linewidth}{!}{
    \begin{tabular}{llcc}
    \hline
    \label{CSS baseline}
         Method&  image backbone& R@1(3D-to-3D) &  R@1(2D-to-3D) \\
    \hline
         TIRG \cite{JGAN}&  ResNet-18& 73.70&  46.60 \\
         HCL \cite{HCL}& ResNet-18& 81.59&  58.65 \\
         GA(TIRG) \cite{huang2022-GA-data-augmentation}& ResNet-18&91.20&- \\
         TIRG+JPM(MSE) \cite{JPM}& ResNet-18& 83.80& -  \\
         TIRG+JPM(Tri) \cite{JPM}& ResNet-18& 83.20& - \\
         LGLI \cite{huang2023-LGLI}& ResNet-18& 93.30& -\\
         MAAF \cite{dodds2020MAAF}& ResNet-50& 87.80& -\\
         CRR \cite{CRR}&  ResNet-101&85.84& -  \\
         JGAN \cite{JGAN}& ResNet-101& 76.07&  48.85\\
         GSCMR \cite{2022-GSCMR}& ResNet-101&81.81& 58.74\\
         TIS \cite{TIS}& Inception-v3&  76.64&  48.02 \\
         LBF(big) \cite{hosseinzadeh2020-locally-LBF}& Faster-RCNN&79.20&  55.69 \\
         LBF(small) \cite{hosseinzadeh2020-locally-LBF}& Faster-RCNN&67.26& 50.31\\
    \hline
    \end{tabular}}
\end{table}

\begin{table}
    \centering
    \caption{Performance comparison on the Shoes \cite{Berg2010-Shoes} dataset.  Notably,  R@K refers to Recall rate for top K. 
A higher value indicates better performance.}
    \resizebox{\linewidth}{!}{
    \begin{tabular}{llcccc}
    \hline
    \label{Shoes baseline}
         Method&  Image encoder& R@1 &  R@10 & R@50 & Average\\
    \hline
         ComposeAE \cite{anwaar2021composeAE}& ResNet-18 & 31.25& 60.30& - &- \\
         TIRG \cite{vo2018TIRG}&  ResNet-50&12.60&  45.45&  69.39& 42.48 \\ 
         VAL($\mathcal{L}_{vv}$) \cite{Chen2020VAL}& ResNet-50& 16.49&  49.12&  73.53& 46.38\\
         VAL($\mathcal{L}_{vv}$ + $\mathcal{L}_{vs}$) \cite{Chen2020VAL}& ResNet-50& 16.98&  49.83&  73.91& 46.91\\
         VAL(GloVe) \cite{Chen2020VAL}& ResNet-50& 17.18&  51.52&  75.83& 48.18\\
         CoSMo \cite{Lee2021CoSMo}&  ResNet-50& 16.72&  48.36&  75.64& 46.91\\
         CLVC-Net \cite{wen2021-CLVC-NET}& ResNet-50& 17.64& 54.39& 79.47&50.50\\
         DCNet \cite{kim2021-DCNet}& ResNet-50&- & 53.82& 79.33&- \\
         SAC w/BERT \cite{jandial2021SAC}& ResNet-50&18.5& 51.73& 77.28& 49.17  \\
         SAC w/Random Emb. \cite{jandial2021SAC}& ResNet-50& 18.11& 52.41& 75.42& 48.64  \\
         ARTEMIS+LSTM \cite{delmas2022ARTEMIS}& ResNet-50 &17.60& 51.05& 76.85& 48.50\\
         ARTEMIS+BiGRU \cite{delmas2022ARTEMIS}& ResNet-50 &18.72& 53.11& 79.31& 50.38\\
         AMC \cite{AMC}& ResNet-50& 19.99& 56.89& 79.27& 52.05  \\
         DATIR \cite{zhao2022-PL4CIR_PLHMQ-twostage}& ResNet-50& 17.20& 51.10& 75.60& 47.97  \\  
         MCR \cite{CRR}& ResNet-50& 17.85& 50.95& 77.24&  48.68 \\ 
         EER \cite{EER}& ResNet-50&20.05& 56.02& 79.94& 52.00  \\ 
         CRN \cite{2023-CRN}& ResNet-50 & 17.19& 53.88& 79.12& 50.06 \\
         Uncertainty \cite{chen2024uncertainty}& ResNet-50 & 18.41& 53.63& 79.84& 50.63 \\
         FashionVLP \cite{Goenka_2022_FashionVLP}& ResNet-50 & - & 49.08& 77.32& -\\
         DWC \cite{huang2023-DWC}& ResNet-50& 18.94& 55.55& 80.19& 51.56\\
         MCEM ($\mathcal{L}_{\small CE}$) \cite{MCEM}& ResNet-50& 15.17& 49.33& 73.78& 46.09\\
         MCEM ($\mathcal{L}_{\small FCE}$) \cite{MCEM}& ResNet-50& 18.13& 54.31& 78.65& 50.36\\
         MCEM($\mathcal{L}_{\small AFCE}$) \cite{MCEM}& ResNet-50& 19.10& 55.37& 79.57& 51.35\\
         AlRet \cite{xu2024-AlRet}& ResNet-50& 18.13& 53.98& 78.81& 50.31\\
         RTIC \cite{shin2021RTIC}& ResNet-50& 43.66& 72.11& - &- \\
         RTIC-GCN \cite{shin2021RTIC}& ResNet-50& 43.38& 72.09&-&-\\
         CRR \cite{CRR}& ResNet-101& 18.41& 56.38& 79.92&  51.57 \\
         CRN \cite{2023-CRN}& Swin & 17.32& 54.15& 79.34& 50.27 \\
         ProVLA \cite{Hu_2023_ICCV-ProVLA}& Swin & 19.20& 56.20& 73.30& 49.57\\
         CRN \cite{2023-CRN}& Swin-L & 18.92& 54.55& 80.04& 51.17 \\
         AlRet \cite{xu2024-AlRet}& CLIP& 21.02& 55.72& 80.77& 52.50\\ 
         PL4CIR \cite{zhao2022-PL4CIR_PLHMQ-twostage}&CLIP-L& 22.88& 58.83& 84.16& 55.29  \\  
         PL4CIR \cite{zhao2022-PL4CIR_PLHMQ-twostage}&CLIP-B& 19.53& 55.65& 80.58& 51.92  \\ 
         TG-CIR \cite{Wen_2023-TG-CIR}& CLIP-B& 25.89& 63.20& 85.07& 58.05  \\     
         DQU-CIR \cite{Wen_2024-DQU-CIR}& CLIP-H& 31.47& 69.19& 88.52& 63.06\\
    \hline
    \end{tabular}}
\end{table}

\begin{table}
    \centering
    \caption{Performance comparison on the CIRR  \cite{liu2021CIRPLANT} dataset.  Notably,  R@K refers to Recall rate for top K. 
A higher value indicates better performance.}

    \resizebox{\linewidth}{!}{
    \begin{tabular}{llcccc}
    \hline
    \label{CIRR baseline}
         Method&  Image encoder& R@1 & R@5 & R@10 & R@50 \\
    \hline
         ComposeAE \cite{shin2021RTIC}& ResNet-18 & - & 29.60& 59.82& - \\
         MCEM($\mathcal{L}_{\small CE}$) \cite{MCEM}& ResNet-18& 14.26& 40.46& 55.61& 85.66\\
         MCEM($\mathcal{L}_{\small FCE}$) \cite{MCEM}& ResNet-18& 16.12& 43.92& 58.87& 86.85\\
         MCEM($\mathcal{L}_{\small AFCE}$) \cite{MCEM}& ResNet-18& 17.48& 46.13& 62.17& 88.91\\
         Ranking-aware \cite{chen2023ranking-aware}& ResNet-50&32.24& 66.63& 79.23& 96.43  \\
         SAC w/BERT \cite{jandial2021SAC}& ResNet-50 & - & 19.56& 45.24&-\\
         SAC w/Random Emb. \cite{jandial2021SAC}& ResNet-50 & - & 20.34& 44.94&- \\
         ARTEMIS+BiGRU \cite{delmas2022ARTEMIS}& ResNet-152 &16.96& 46.10& 61.31& 87.73\\
         CIRPLANT \cite{liu2021CIRPLANT}& ResNet-152&15.18& 43.36& 60.48& 87.64 \\
         CIRPLANT w/ OSCAR \cite{liu2021CIRPLANT}& ResNet-152& 19.55& 52.55& 68.39& 92.38\\
         CASE \cite{levy2023CASE}& ViT& 48.00& 79.11& 87.25& 97.57\\
         ComqueryFormer \cite{ComqueryFormer}& Swin & 25.76& 61.76& 75.90& 95.13\\ 
         CLIP4CIR \cite{baldrati2022-CLIP4CIR}&CLIP& 38.53& 69.98& 81.86& 95.93   \\
         CLIP4CIR3 \cite{CLIP4CIR3}&CLIP& 44.82& 77.04& 86.65& 97.90   \\
         SPN(TG-CIR) \cite{feng2024data_generation-SPN}& CLIP&47.28& 79.13& 87.98&97.54\\
         SPN(CLIP4CIR) \cite{feng2024data_generation-SPN}& CLIP& 45.33& 78.07& 87.61& 98.17\\
         Combiner \cite{baldrati2022combiner}& CLIP&33.59& 65.35& 77.35& 95.21  \\
         MCEM($\mathcal{L}_{\small AFCE}$) \cite{MCEM}& CLIP& 39.80& 74.24& 85.71& 97.23\\
         TG-CIR \cite{Wen_2023-TG-CIR}& CLIP-B& 45.25& 78.29& 87.16& 97.30   \\
         CIReVL \cite{karthik2024-CIReVL}& CLIP-B& 23.94& 52.51& 66.0& 86.95 \\
         SEARLE-OTI \cite{Baldrati2023SEARLE}& CLIP-B& 24.27& 53.25& 66.10& 88.84\\
         SEARLE \cite{Baldrati2023SEARLE}& CLIP-B& 24.00& 53.42& 66.82& 89.78\\
         PLI \cite{chen2023-PLI}& CLIP-B& 18.80& 46.07& 60.75& 86.41\\
         SEARLE-XL \cite{Baldrati2023SEARLE}& CLIP-L & 24.24& 52.48& 66.29& 88.84\\
         SEARLE-XL-OTI \cite{Baldrati2023SEARLE}& CLIP-L & 24.87& 52.31& 66.29& 88.58\\
         CIReVL \cite{karthik2024-CIReVL}& CLIP-L& 24.55& 52.31& 64.92& 86.34\\
         Context-I2W \cite{tang2023contexti2w}& CLIP-L & 25.6& 55.1 & 68.5& 89.8 \\
         Pic2Word \cite{saito2023pic2word}& CLIP-L& 23.90& 51.70& 65.30& 87.80 \\
         CompoDiff(with SynthTriplets18M) \cite{gu2024compodiff}&  CLIP-L&  18.24&  53.14&  70.82&  90.25\\
         LinCIR \cite{gu2024LinCIR}& CLIP-L& 25.04&  53.25& 66.68& - \\ 
         PLI \cite{chen2023-PLI}& CLIP L& 25.52& 54.58& 67.59& 88.70\\
         KEDs \cite{suo2024KEDs}& CLIP-L& 26.4& 54.8& 67.2& 89.2\\
         CIReVL \cite{karthik2024-CIReVL}& CLIP-G& 34.65& 64.29& 75.06& 91.66\\
         LinCIR \cite{gu2024LinCIR}& CLIP-G& 35.25&  64.72& 76.05& - \\
         CompoDiff(with SynthTriplets18M) \cite{gu2024compodiff}&  CLIP-G&  26.71&  55.14&  74.52&  92.01\\
         LinCIR \cite{gu2024LinCIR}& CLIP-H& 33.83&  63.52& 75.35& - \\  
         DQU-CIR \cite{Wen_2024-DQU-CIR}& CLIP-H& 46.22& 78.17& 87.64& 97.81\\
         PLI \cite{chen2023-PLI}& BLIP& 27.23& 58.87& 71.40& 91.25\\
         BLIP4CIR2 \cite{liu2024-BLIP4CIR2}& BLIP& 40.17&71.81& 83.18& 95.69\\
         BLIP4CIR2+Bi \cite{liu2024-BLIP4CIR2}& BLIP& 40.15&73.08& 83.88& 96.27\\
         SPN(BLIP4CIR1) \cite{feng2024data_generation-SPN}& BLIP& 46.43& 77.64& 87.01& 97.06\\
         SPN(SPRC) \cite{feng2024data_generation-SPN}& BLIP-2& 55.06& 83.83& 90.87& 98.29\\
         BLIP4CIR1 \cite{liu2023BLIP4CIR1}& BLIP-B&46.83& 78.59& 88.04& 97.08\\
    \hline
    \end{tabular}}
\end{table}
    
In  \cite{saito2023pic2word},  zero-shot methods operate without the supervision of large-scale triplets and can be applied in an open domain. 
In  \cite{saito2023pic2word},  the Pic2Word model,  which maps images to word tokens,  is trained on large-scale image-caption pairs and unlabeled images. 
Requiring even less data,  SEARLE  \cite{Baldrati2023SEARLE} uses GPT-powered regularization to generate pseudo-word tokens. Following vision-by-language paradigm,  CIReVL \cite{karthik2024-CIReVL} uses language as an abstraction layer for reasoning about visual content. PALAVRE \cite{cohen2022-PALAVRA} is a textual inversion-based two-stage approach with a pre-trained mapping function and a subsequent optimization of the pseudo-word token. Compared to Pic2Word \cite{saito2023pic2word},  SEARLE employs a smaller dataset and a more complex combination of loss functions.
Both the two projection-based ZS-CIR methods convert the whole information of the image into the same pseudo-word,  which limits the flexibility of adaptively select information. 
To improve this,  contexti2w \cite{tang2023contexti2w} is a context-dependent mapping network to adaptively convert description-relevant image information into a pseudo word in a hierarchical mode. 
 \cite{tang2023contexti2w} considers manipulation descriptions and learnable queries to be multi-level constraints for visual information filtering. 
An interactive image retrieval system is proposed in Enahancing \cite{Zhu_2024-enhancing},  which composes text feedback with former image based on InstructeBLIP \cite{dai2023instructblip},  a pre-trained VLM-based image captioner,  which uses the vicuna-7b \cite{zheng2024-vicuna-7b} to generate captions and a ViT-g/14 \cite{dosovitskiy2021-viT-g/14} model to extract image features. 

In essence,  zero-shot retrieval involves text inversion,  whereby the reference image is fed into an image encoder and then translated into text. This text is then amalgamated with the provided textual input,  facilitating the retrieval of target image. Leveraging this textual inversion process,  the task of composite image retrieval is degenerated into standard text-to-image retrieval.

\subsubsection{Hybrid Methods}\label{Hybrid methods}
To leverage the strengths of various methods,  some studies propose to integrate multiple approaches (e.g.,  CNN and CLIP). 
The Language-Guided Local Infiltration (LGLI) system  \cite{huang2023-LGLI} aims to improve the integration of textual and image features. It includes a Language Prompt Visual Localization (LPVL) module that generates masks to accurately identify the semantic areas associated with the modification text,  and then employs a Text Infiltration with Local Awareness (TILA) module to fine-tune the reference image,  resulting in an output that seamlessly merges image and text.
The Dynamic Weighted Combiner (DWC)  \cite{huang2023-DWC} further addresses the challenge and offers three key benefits. First,  it features an Editable Modality De-equalizer (EMD) to balance different contributions of various modalities and incorporate two modality feature editors and an adaptive weighted combiner. Second,  to minimize labeling noise and data bias,  it introduces a dynamic soft-similarity label generator (SSG) that enhances noisy supervision. Last,  it presents a CLIP-based mutual enhancement module 
to bridge the gap between modalities. 

\subsubsection{Summary}
    In conclusion,  image-text composite retrieval methods typically utilize various architectures,  including traditional CNN-based,  transformer-based,  large model-based,  and hybrid methods. The field is rapidly evolving,  and several promising future directions are proposed. 
\begin{enumerate}
    \item Improving Model Ability on Bridging Modality Gap: Future work could focus on bridging modality gap by 
     developing techniques to align visual and textual features more effectively. This ensures that the models can seamlessly interpret and integrate information from both modalities.
    \item Handling Open-Domain Scenarios: To make image-text composite retrieval systems more versatile,  it is crucial to enable them to operate on open-domain data. This requires designing models that generalize well across various topics,  styles,  and contexts without being restricted to specific datasets. Techniques such as domain adaptation,  transfer learning,  and zero-shot learning could play significant roles in achieving this goal.
    \item Retrieval with Fewer Data or Weak Supervision: The reliance on large annotated datasets is a significant bottleneck. Future approaches could explore methods to reduce data dependency through semi-supervised,  weakly supervised,  and unsupervised learning strategies. Utilizing synthetic data generation,  self-supervised learning,  and leveraging external knowledge bases can also help models learn effectively from fewer labeled examples. 
\end{enumerate}

\subsection{Other Mutimodal Composite Retrieval}\label{sec4}
Most of multimodal composite retrieval methods focus on the composition of visual and language modalities. As shown in Table \ref{table methods OMCR},  there are still several studies exploring diverse modalities such as sketches,  audio,  motion,  etc. 

\subsubsection{Composition of Other Modalities}
The exploration of compositional modalities generally includes image retrieval  \cite{mou2023-T2I-Adapter,  2020-DGMFE,  2013-NovaMedSearch,  zhang2024fabric} and document retrieval  \cite{2013-document-retrieval,  DocStruct4M}. The T2I-Adapter  \cite{mou2023-T2I-Adapter} aligns internal knowledge in text-to-image (T2I) models with external control signals such as sketches,  key poses,  and color maps. This integration facilitates the combination of text with other modalities in the diffusion-based generation process,  demonstrating notable composability and generalization. Similarly,  MMFR  \cite{2022-MMFR} composes audio feature representations with text information to perform image retrieval. 
It first converts the original audio input to text using VGGISH  \cite{hershey2017-VGGISH},  pretrained on Audioset  \cite{2017-audioset}. A feature fusion module then combines the transformed audio representation with text,  enhancing the semantic distinction of pronunciation-based audio features and bridging the heterogeneous gap.
    
\subsubsection{Tri-modal Composition}
Some methods integrate information from three distinct modalities. For instance,  SIMC  \cite{2003-SIMC} demonstrates that combining data from visual,  audio,  speech,  video and text modalities significantly enhances semantic labeling performance. This is achieved by first classifying concepts based on individual modalities and then integrating them.
LAVIMO  \cite{yin2024-LAVIMO} is a Unified Language-Video-Motion Alignment framework. It employs three encoders,  each initialized with pre-trained models to extract features from motion,  text,  and video,  respectively. These modalities are then aligned towards a joint embedding space through a custom attention mechanism.
TriCoLo  \cite{ruan2023-TriCoLo} combines information from text,  multi-view images,  and 3D voxels to learn a shared embedding,  utilizing contrastive learning for effective text-to-shape retrieval.
TWPW  \cite{changpinyo2021-TWPW} is an image retrieval framework that employs both text and mouse traces as queries. It constructs a base image retrieval model by leveraging image-text matching data from the Localized Narratives dataset  \cite{ponttuset2020-Localized-Narratives},  subsequently incorporating bounding boxes derived from mouse trace segments.

\subsubsection{Summary} 
In the study of multimodal composite retrieval,  significant progress has been made beyond traditional visual and language forms. Research has broadened to encompass various types,  including sketches,  audio,  and motion,  etc. demonstrating creative methods for merging and aligning these different data forms. By integrating retrieval techniques,  there is optimism for developing a universal multimodal system that can achieve enhanced granularity,  composability,  and generalization.
Future research could focus on developing more efficient methods for integrating a wider range of modalities. This includes exploring novel combinations of existing modalities and incorporating emerging types such as sensor data or haptic feedback.

\section{Benchmarks and Experiments}\label{benchmark}

\subsection{Datasets} 
Table \ref{datasets} organizes frequently used benchmarks for multimodal composite retrieval,  containing three primary tasks: image-text composite editing,  image-text composite retrieval,  and other multimodal composite retrieval. In order to facilitate the research, their detailed information and download links have been provided.

\subsection{Experimental Results} 
For having an in-depth insight on the results,  We provide the performance comparison among a large number of image-text composite retrieval methods across various datasets,  including Fashion-IQ (Table \ref{Fashion-IQ baseline} and Table \ref{Fashion-IQ baseline 2}),  Fashion200k (Table \ref{Fashion200k baseline}),  MIT-States (Table \ref{MIT-States baseline}),  CSS (Table \ref{CSS baseline}),  Shoes (Table \ref{Shoes baseline}),  and CIRR (Table \ref{CIRR baseline}). 
Regarding the Fashion-IQ baseline,  we report the performance using both the VAL split and original split. Compared to the original split,  VAL constructs smaller candidate sets by merging the reference and target images,  which decreases the number of test images and slightly improves performance across all models. For a fair comparison,  we present results on both the VAL and original split for Fashion-IQ dataset.
The evaluation metrics are outlined in the respective table captions. The results indicate that Transformer-based and VLP-based methods outperform traditional CNN-based approaches. This,  to a large extent,  benefits from the introduction of self-attention mechanism and larger datasets. This also inspire us for future prospective to advance this field.

\section{Discussion and Future Directions} \label{discussion}

The aforementioned sections have delved into research on multimodal composite retrieval. Despite significant strides,  there still exist several challenges and open questions.  In this section,  we summarize key challenges and offer discussions on potential future directions.

\textbf{Bridging Modality Gaps.} A fundamental challenge lies in effectively integrating diverse modalities. Existing techniques such as attention mechanisms,  graph-based approaches,  and other general-purpose methods have been employed to refine the integration of multimodal information,  making it more nuanced and holistic. However,  achieving a unified understanding across multiple modalities remains an ongoing challenge. More advanced alignment techniques may be explored to seamlessly integrate diverse modalities.

\textbf{Robustness and Generalization.} Credibility and reliability in real-world applications is essential. Deep neural networks (DNNs) are known to be susceptible to adversarial attacks,  yet the adversarial robustness of multimodal composite retrieval systems has received less attention. Recent advancements in enhancing adversarial robustness and generalization  \cite{huang2022-GA-data-augmentation} have primarily focused on improving generalization through adversarial and isotropic gradient augmentation. Therefore,  adversarial robustness and generalization of multimodal models should be paid more efforts in future research across diverse datasets and scenarios. 

\textbf{Scalability and Flexibility.} As datasets grow in size and complexity,  the scalability of retrieval systems becomes increasingly important. Leveraging large-scale pre-trained models,  particularly the text-processing capabilities of large language models (LLMs)  \cite{brown2020-GPT-3, devlin2019BERT},  offers a promising opportunity to enhance the generation and retrieval of information across modalities.

\textbf{Universal Multimodal Composite Retrieval.} Existing methods typically focus on various combinations of different modalities,  such as combination of image and text or other three modalities. However,  there is significant value in exploring generalized approaches that can integrate a wider range of modalities. Exploring a universal retrieval system across more modalities,  including but not limited to image,  text,  audio,  video,  is a promising direction.

\textbf{Interpretability.} Most current models are based on deep learning,  which operate as enigmatic \textit{black boxes}.  Enhancing interpretability is crucial for understanding decision-making processes and improving user trust in multimodal composite retireval systems. It is still a long-standing work to explore  how models make decisions.

\section{Conclusion}\label{conclusion}
This survey has explored the evolving field of multimodal composite retrieval,  which combines multiple modalities,  such as text,  images,  and audio,  in order to improve retrieval accuracy and user interaction. We reviewed over more than 200 advanced methodologies,  organizing them into three main categories: image-text composite editing,  image-text composite retrieval,  and other multimodal composite retrieval. This taxonomy clarifies the current research landscape and highlights the strengths and weaknesses of existing approaches. Moreover,  we identified key challenges and proposed future research directions to foster innovation in this area.  Our survey serves as a valuable resource for researchers and practitioners,  offering insights into the current state of multimodal composite retrieval and its potential for further advancement.

\bibliographystyle{IEEEtranS}
\bibliography{ref}

\ifCLASSOPTIONcaptionsoff
  \newpage
\fi



%
%

\begin{IEEEbiography}[{\includegraphics[width=1in, height=1.25in, clip, keepaspectratio]{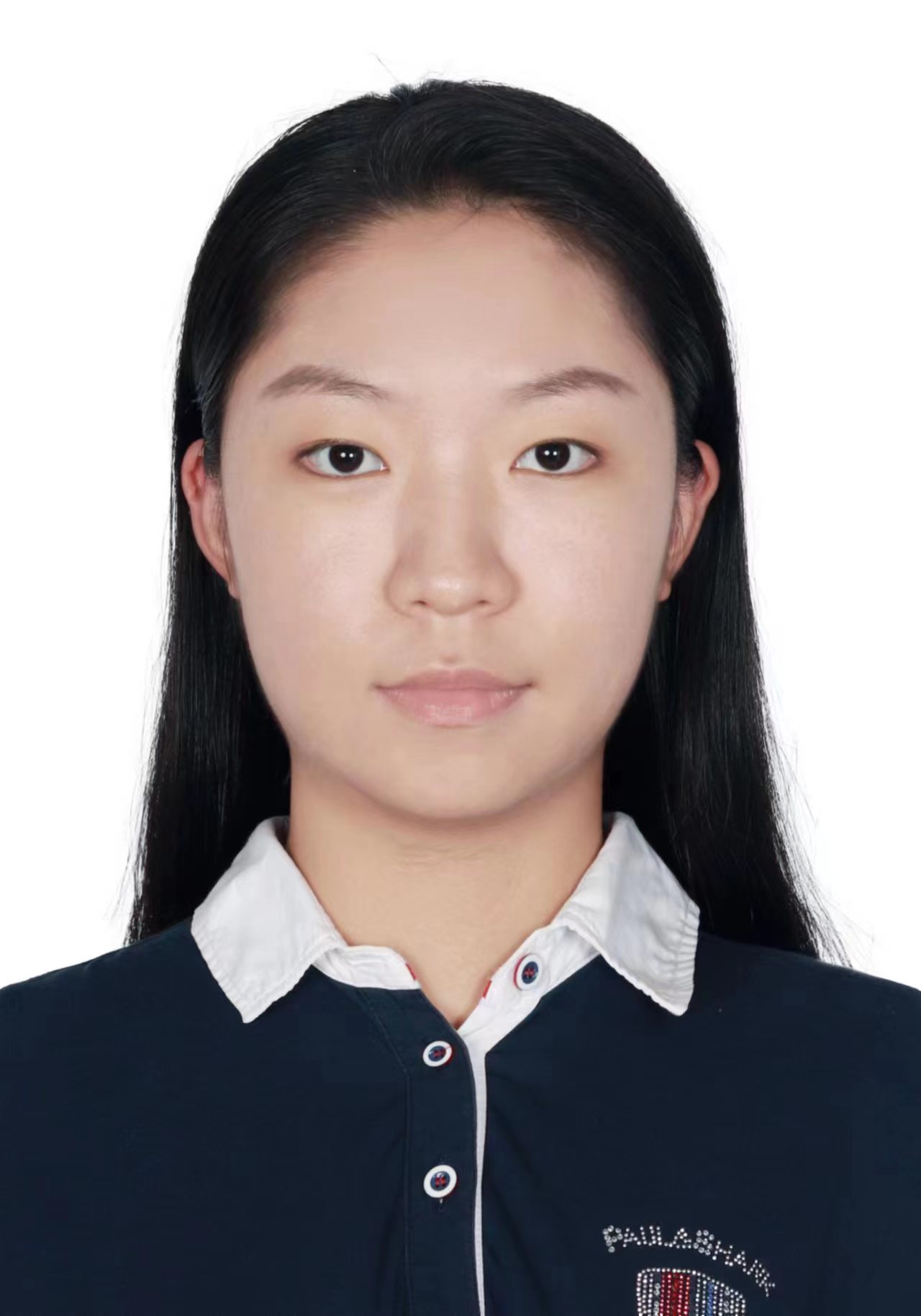}}]
{Suyan Li} is now pursuing her B.S. degree at Chongqing University,  Chongqing,  China. Her research interests include machine learning and multimodal retrieval.
\end{IEEEbiography}

\begin{IEEEbiography}[{\includegraphics[width=1in, height=1.25in, clip, keepaspectratio]{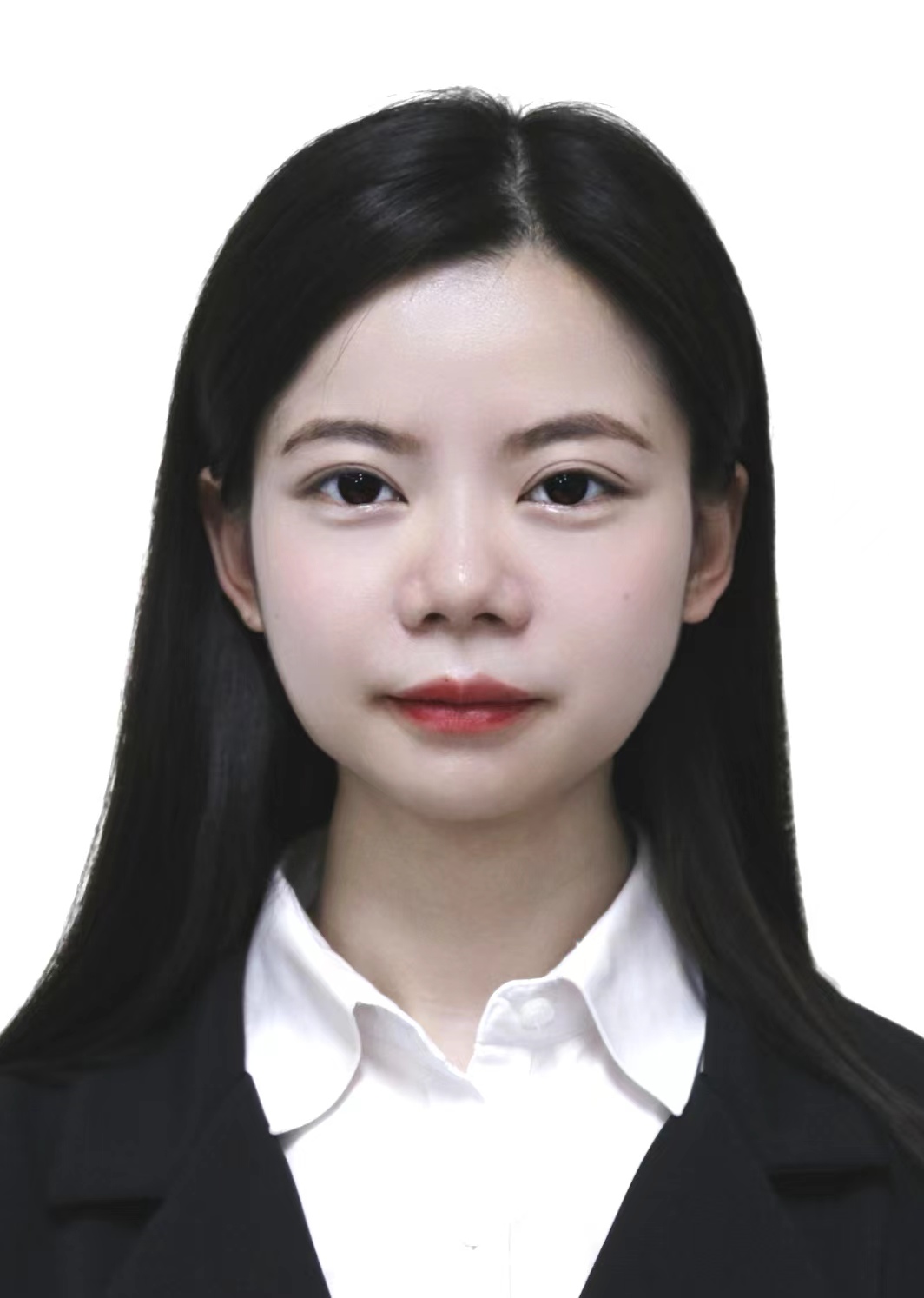}}]
{Fuxiang Huang} received her Ph.D degree in Information and Communication Engineering from Chongqing University,  Chongqing,  China,  in 2023 and is currently working as a postdoc at The Hong Kong University of Science and Technology. She has published more than 10 technical articles in top journals and conferences,  such as IEEE T-PAMI,  IJCV,  T-IP,  T-NNLS,  T-MM,  T-CSVT,  CVPR and AAAI. 
Her current research interests include computer vision,  deep learning,  domain adaptation and multi-modal retrieval.
\end{IEEEbiography}

\begin{IEEEbiography}[{\includegraphics[width=1in, height=1.25in, clip, keepaspectratio]{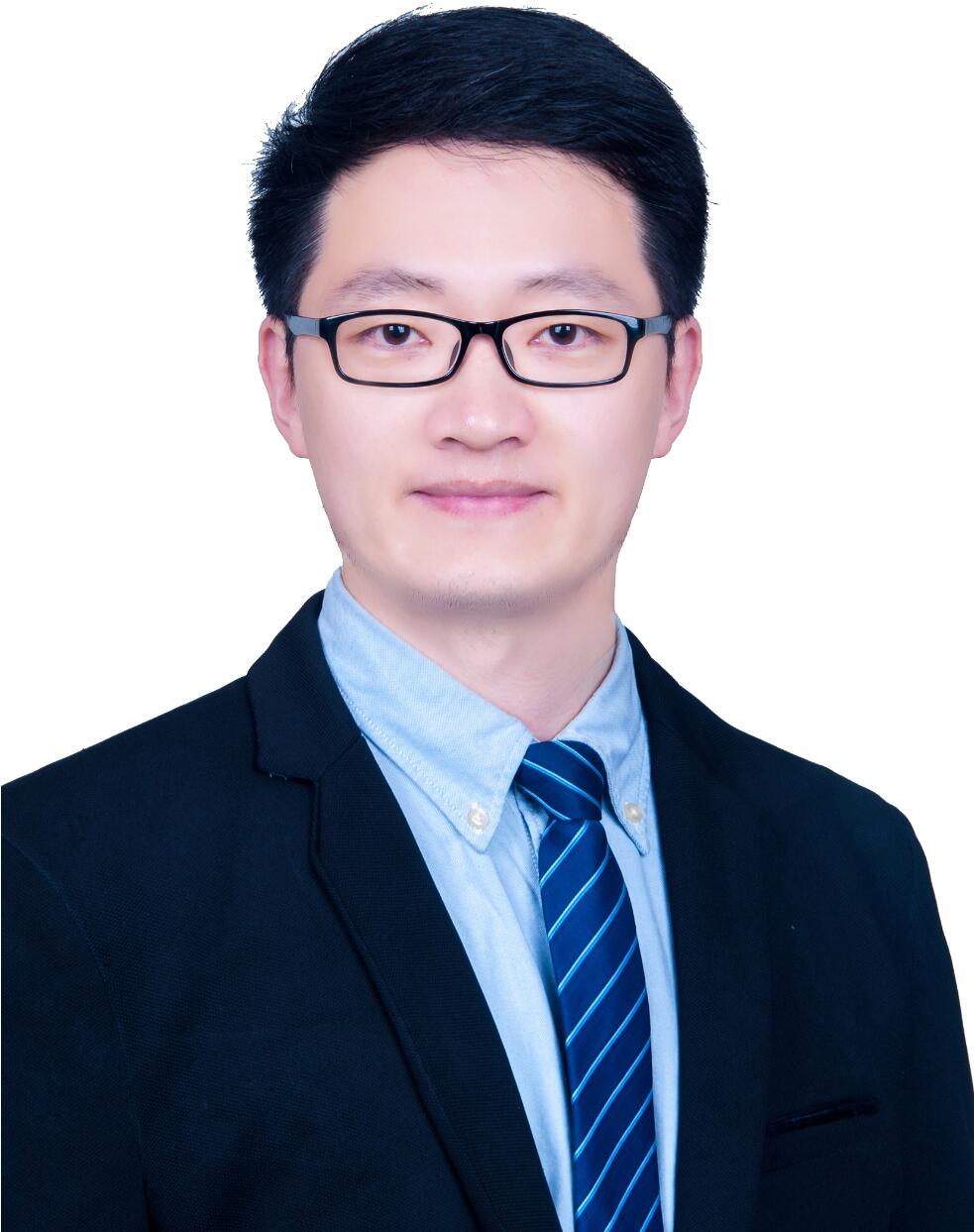}}] {Lei Zhang} received his Ph.D degree in Circuits and Systems from the College of Communication Engineering,  Chongqing University,  Chongqing,  China,  in 2013. He worked as a Post-Doctoral Fellow with The Hong Kong Polytechnic University,  Hong Kong,  from 2013 to 2015. He is currently a Full Professor with Chongqing University and the director of the Chongqing Key Laboratory of Bio-perception and Multimodal Intelligent Information Processing. He has authored more than 150 scientific papers in top journals and conferences,  including IEEE T-PAMI,  IJCV,  T-IP,  T-MM,  T-CSVT,  T-NNLS,  CVPR,  ICCV,  ECCV,  ICML,  AAAI,  ACM MM,  IJCAI,  etc. He is on the Editorial Boards of several journals,  such as IEEE Transactions on Instrumentation and Measurement,  Neural Networks (Elsevier),  etc. Dr. Zhang was a recipient of the 2019 ACM SIGAI Rising Star Award.

His current research interests include deep learning,  transfer learning,  domain adaptation and computer vision. He is a Senior Member of IEEE.
\end{IEEEbiography}







\end{document}